\documentclass{IEEEtran}

\usepackage{subfigure}
\usepackage{cite}
\usepackage{amsmath,amssymb,amsfonts}
\usepackage[ruled,linesnumbered]{algorithm2e}

\usepackage{color}
\usepackage{graphicx}
\usepackage{textcomp}
\usepackage{booktabs}
\usepackage{multirow}
\usepackage{bbding}
\usepackage{bm}

\def\BibTeX{{\rm B\kern-.05em{\sc i\kern-.025em b}\kern-.08em
	T\kern-.1667em\lower.7ex\hbox{E}\kern-.125emX}}

\begin{document}
\title{AIGC for Industrial Time Series: From Deep Generative Models to Large Generative Models }
\author{
{
    Lei Ren, \IEEEmembership{Member, IEEE,}
    Haiteng Wang, \IEEEmembership{Student Member, IEEE,}
    Jinwang Li, \IEEEmembership{Student Member, IEEE,}
    
    Yang Tang, \IEEEmembership{Fellow, IEEE,}  
    and Chunhua Yang, \IEEEmembership{Fellow, IEEE}
}
\thanks{	
    {
    	This work has been submitted to the IEEE for possible publication. Copyright may be transferred without notice, after which this version may no longer be accessible.
    	
Lei Ren, Haiteng Wang and Jinwang Li are with School of Automation Science and Electrical Engineering, Beihang University, Beijing 100191, China. (email: renlei@buaa.edu.cn; wanghaiteng@buaa.edu.cn; Jinwangli@buaa.edu.cn).

Yang Tang is with the Key Laboratory of Smart Manufacturing in Energy Chemical Process, Ministry of Education, East China University of Science and Technology, Shanghai 200237, China (e-mail: yangtang@ecust.edu.cn).

Chunhua Yang is with the School of Automation, Central South University, Changsha 410083, China. (e-mail:  ychh@csu.edu.cn).
    }
}
}
\maketitle
\begin{abstract}

With the remarkable success of generative models like ChatGPT, Artificial Intelligence Generated Content (AIGC) is undergoing explosive development. Not limited to text and images, generative models can generate industrial time series data, addressing challenges such as the difficulty of data collection and data annotation. Due to their outstanding generation ability, they have been widely used in Internet of Things, metaverse, and cyber-physical-social systems to enhance the efficiency of industrial production. In this paper, we present a comprehensive overview of generative models for industrial time series from deep generative models (DGMs) to large generative models (LGMs). First, a DGM-based AIGC framework is proposed for industrial time series generation.  Within this framework, we survey advanced industrial DGMs and present a multi-perspective categorization.  Then, we systematically propose the roadmap to construct industrial LGMs from four aspects: large-scale industrial dataset, LGMs architecture for complex industrial characteristics, self-supervised training for industrial time series, and fine-tuning of industrial downstream tasks. Furthermore, we introduce an evaluation benchmark that systematically assesses fidelity, diversity, and utility. We include a case study on aircraft engine maintenance, demonstrating the application of DGMs in industrial predictive maintenance.
Finally, we conclude the challenges and future directions to enable the development of generative models in industry.  
	
\end{abstract}


\section{Introduction}
\label{sec:introduction}

Industry 5.0 incorporates cyber-physical-social elements into manufacturing, emphasizing digital-physical interaction and human-machine collaboration, effectively connecting the Internet of devices, things, and people. Currently, with the development of Artificial Intelligence Generated Content (AIGC), metaverse and digital twins technologies, industrial big data can be used to create digital manufacturing and industrial processes \cite{yang2024generative}. It facilitates significant growth in productivity, efficiency, and effectiveness in Industry 5.0 and cyber-physical-social systems (CPSS)\cite{wang2023smart}.

Industrial time series data, including equipment sensor data, production-line operation data, and system status data, is an important foundation for a broad range of industrial intelligence applications\cite{ren2023deep}. Currently, more and more industrial intelligence applications, such as intelligent control systems\cite{ding2020secure,tang2023gru}, predictive maintenance\cite{yin2020anomaly,pei2022bayesian}, and fault diagnosis\cite{liu2023lightweight,zheng2023interval}, rely on mining and analyzing industrial time series. In particular, the success of industrial temporal deep learning methods is also dependent on learning and extracting features from industrial time series data.

The diversity, quality, and quantity of industrial time series are critical for improving system performance and facilitating industrial intelligence applications. Unfortunately, in industrial scenarios, obtaining such data can be challenging due to the difficulty of industrial data collection, data annotation, and privacy concerns. For example, data collection for industrial engine equipment requires significant labor costs. Fatigue testing may even damage industrial equipment. Moreover, data silos and sensitive information about industrial data further hinder the creation of rich and diverse datasets. To address these challenges, data generation through generative models provides a promising solution. 

In the past two years, large generative models, such as DALL·E 2\cite{110-ramesh2022hierarchical} and ChatGPT\cite{109-ouyang2022training} have ushered in a new era of AIGC. These models can generate new data with similar statistical characteristics by learning the distribution of existing data. This enable significant progress in the fields of image synthesis, text generation, and molecular design, among other tasks. Microsoft, OpenAI, and others have used synthetic data from generative models to train intelligent models. This shows the powerful capability of generative models to deal with challenges related to data annotation and collection. Generative models can be classified into large generative models (LGMs) and deep generative models (DGMs). DGMs for industrial data generation, include generative adversarial networks (GAN)\cite{101-jiang2019gan,102-klopries2024itf,103-li2019mad,104-goodfellow2014generative}, variational autoencoders (VAE)\cite{107-kingma2013auto, 21-Zha2022NormalizedConditionala,30-Jia2022ImprovingPerformancea}, and diffusion 
models\cite{106-ho2020denoising,46-yan2023bearing,52-dai2023timeddpm}. Unlike traditional data extension methods such as time-shifting, downsampling, and panning, DGMs are able to learn and model complex distributions of complex data to generate diverse new samples.
 
In the industrial field, DGMs have become a research hotspot and have been widely applied. Its core tasks can be roughly categorized into five: limited sample augmentation\cite{7-gao2022deep,8-zhang2022novel,9-huang2023hybrid}, imbalanced data synthesis\cite{22-Pen2022NovelBearing,23-Son2023DeepGenerative}, sensor signal imputation\cite{6-xiao2023imputation,11-velasco2022analysis}, sensor signal denoising\cite{24-Liu2023HarmonicReducerb,27-Wei2023MultisensorSignalsa}, privacy protection\cite{37-Pal2022APROSecreta,38-Kes2020PrivacyPreservingFrameworkBasedBlockchain,41-chen2021private}. DGMs help to alleviate issues such as data privacy and shortage. By training generative models, companies can synthesize data without disclosing sensitive information, thus protecting data privacy. In addition, data generation is able to expand limited real datasets, reduce the cost of data acquisition, and improve the generality of models. Generating samples with less variety solves the problem of data imbalance. Recent research has exploited the ability of DGMs to generate real data samples for industrial applications such as anomaly detection\cite{1-zhang2022anomaly,2-fan2023effective}, multi-classification fault instance generation (to address data imbalance issues)\cite{33-Ort2020DataAugmentationb,34-Gao2020DataAugmentationb}, trust boundary protection\cite{hassan2020robust,hassan2020adaptive}, and platform monitoring\cite{nunez2019neural,kabugo2019process}.

\begin{figure}[!tpb]
    \centering
    \includegraphics[width=\columnwidth]{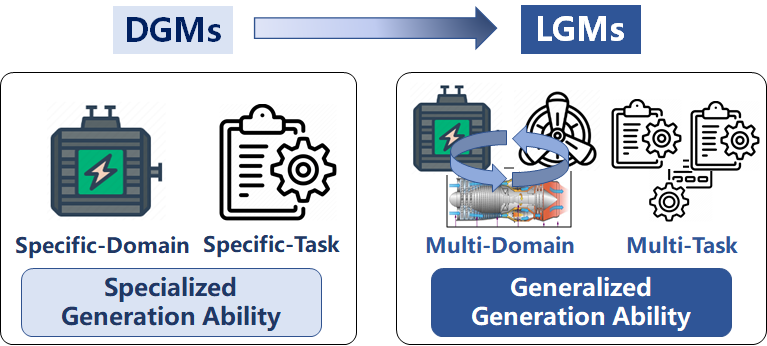}
    \caption{The comparison between DGMs and LGMs.}
    \label{fig:DGMs-LGMs}
\end{figure}

Since generative models have been applied in various fields and have shown promising results, there have been recent studies analyzing and surveying DGMs.  Comprehensive reviews, such as the survey\cite{yang2023diffusion}, provide profound insights into diffusion models across domains such as image synthesis, time series prediction, and natural language generation. Additionally, domain-specific reviews like \cite{de2022deep} presented a review about the application of DGMs in the industry. Similarly, with the rapid development of diffusion models, there are some reviews on diffusion models in specific fields such as biological signals\cite{kazerouni2023diffusion} or computer vision\cite{croitoru2023diffusion}. Due to the significance of time series in industry, literature \cite{ren2023deep} has surveyed time series data modeling issues.
In addition to DGMs with specific-domain, specific-task, and specified generation ability, LGMs with multi-domain, multi-task, and generalized generation ability have made significant progress, as shown in Fig. \ref{fig:DGMs-LGMs}. Some recent studies\cite{cao2024survey, cao2023comprehensive} have investigated the  LGMs in general applications.

Although these existing works review deep generative models, GAN-based methods, and deep learning-based industrial applications, we have identified a gap in the literature regarding the classification and application of generative models in industrial time series.  As time series data stands as the most prevalent and essential data type in industry, its generation holds significant importance. Moreover, there is still a lack of sufficient research on how to construct LGMs in industry. Therefore, this survey focuses on a comprehensive review of DGMs in industrial time series and also analyzes the critical technologies required to construct industrial LGMs in the industrial field. The comparison between existing related surveys and this paper is explicitly shown in Table \ref{tab:Comparison-survey}. The contribution of the paper can be summarized as follows:

1) This paper provides a comprehensive overview of the generative models in the industry, from the current state-of-the-art DGMs to the future promising LGMs.

2) A DGM-based AIGC framework is proposed for industrial time series generation. Within the framework, this paper surveys advanced industrial DGMs and presents a multi-perspective categorization.

3) This paper proposes a systematic roadmap of the critical technologies required to construct industrial LGMs that are suitable for industrial time series data.

4) This paper establishes a comprehensive evaluation benchmark for generative models in industrial time series, systematically assessing fidelity, diversity, and utility. Additionally, a case study on aircraft engine maintenance is presented, demonstrating the effectiveness of DGMs in industrial predictive maintenance.

\begin{table}[!tpb]
\centering
\caption{Comparison with existing relevant surveys}
\label{tab:Comparison-survey}
\begin{tabular}{ccccc}
\hline
Refs                & Time Series & DGMs & LGMs & Industrial Applications   \\ \hline
\cite{croitoru2023diffusion}            &             & \checkmark    &      & Computer vision      \\
\cite{cao2024survey}           &             & \checkmark    &  \checkmark  &  General applications \\
\cite{gao2022generative}        & \checkmark            & \checkmark      &      & Spatio-temporal data \\
\cite{qiu2023large}            &             & \checkmark    &    &  Health informatics               \\

\cite{de2022deep}            &             & \checkmark    &     &  \checkmark                   \\
\cite{ren2023deep}                 & \checkmark           &      &     &  \checkmark                    \\
Proposed            & \checkmark           & \checkmark    & \checkmark    &  \checkmark                    \\ \hline
\end{tabular}
\end{table}

\section{Deep Generative Models}
In this section, we introduce general DGMs architectures and compare their characteristics and applications.

\begin{figure*}
    \centering
    \includegraphics[width=2\columnwidth]{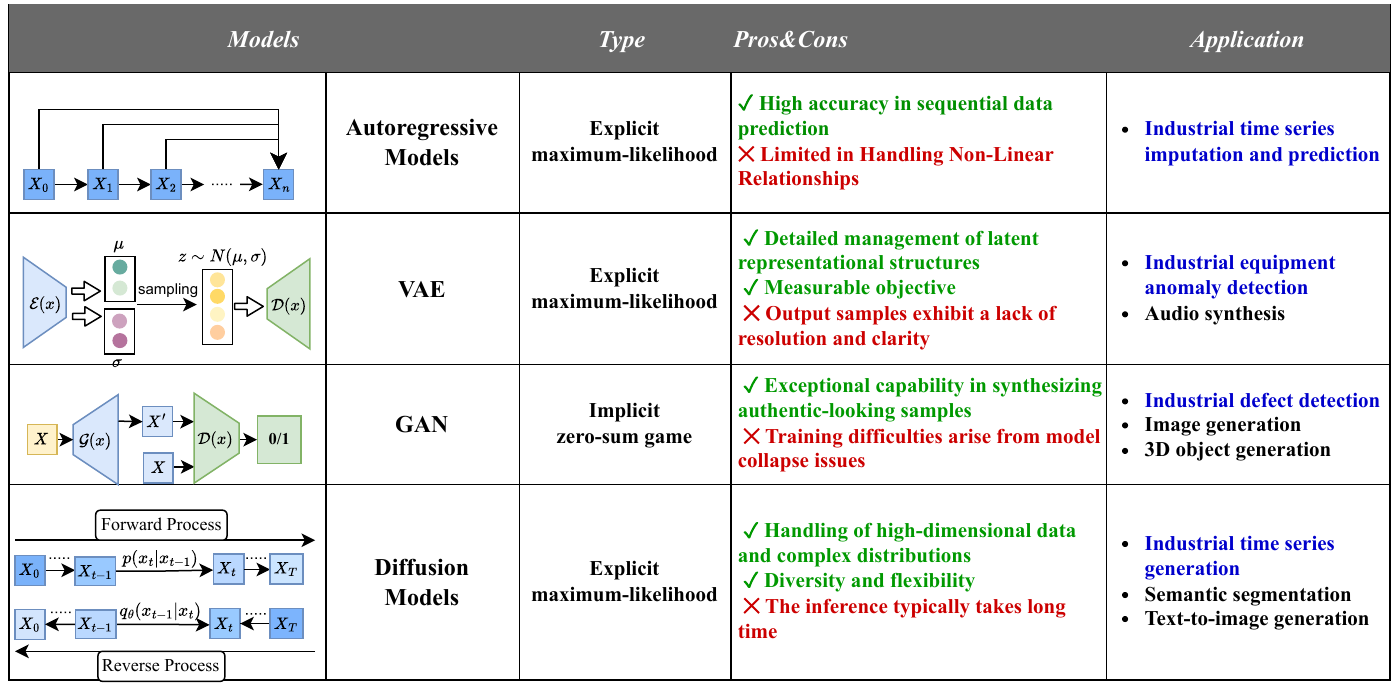}
    \caption{The comparison of mainstream DGMs and their introduction.}
    \label{fig:vae}
\end{figure*}

\subsection{Autoregressive Model}
An autoregressive model (AR) \cite{65-akaike1969fitting} predicts future values in a time series based on past observations. The autoregressive model of order \( p \) (often termed \(AR(p)\) model) is defined as:
\begin{equation}
    X_t = c + \phi_1 X_{t-1} + \phi_2 X_{t-2} + \ldots + \phi_p X_{t-p} + \epsilon_t
\end{equation}
where \( X_t \) is the value of the series at time \( t \), \( c \) is a constant, \( \phi_1, \phi_2, \ldots, \phi_p \) are the model parameters, \( \epsilon_t \) is white noise. For Gaussian noise, the negative log-likelihood serves as the loss function:
\begin{equation}
    \mathcal{L} = -\sum_{t=p+1}^{T} \log f(x_t | x_{t-1}, x_{t-2}, \ldots, x_{t-p})
\end{equation}
Minimizing this function yields the maximum likelihood estimates of the parameters.

AR models are widely used in industrial forecasting. Wang et al. \cite{wang2020forecasting} introduced a hybrid model combining nonlinear autoregressive neural networks and empirical mode decomposition to improve industrial iron ore import volume predictions.

\subsection{Variational Autoencoder (VAE)}

A Variational Autoencoder (VAE) extends the traditional autoencoder (AE) by adopting a probabilistic approach to model data generation. Instead of directly encoding and reconstructing data, VAEs \cite{107-kingma2013auto} map inputs to a latent distribution, from which new samples can be generated. The loss function consists of a reconstruction loss and a regularization term, encouraging the latent space to approximate a standard normal distribution.

Given observed data $x$ and latent variable $z$, the VAE loss is:
\begin{equation}
    \mathcal{L}_{\text{VAE}} (\phi,\theta) = -\mathbb{E}_{q_{\phi}(z|x)}[\log p_{\theta}(x|z)] + D_{\text{KL}}(q_{\phi}(z|x) || p(z))
\end{equation}
Where $p(z)$ is the Prior distribution, typically $N(z; 0, I)$, $q_{\phi}(z|x)$is the Encoder output, approximating the posterior distribution, and $p_{\theta}(x|z)$ is the Decoder output, reconstructing data from $z$.

While VAE has gained significant academic interest, its adoption in industry requires technological maturity and stability. It has been applied to anomaly detection in industrial settings \cite{71-suhanomaly,72-suh2016echo}.

\subsection{Generative Adversarial Network (GAN)}

Generative Adversarial Networks (GANs) were first introduced by Goodfellow et al. \cite{104-goodfellow2014generative} in 2014 as an unsupervised learning algorithm based on a zero-sum game between two neural networks: a generator (G) and a discriminator (D). Unlike traditional methods, GANs do not require prior knowledge or assumptions about the data distribution.
Given real data  $x$  and random noise  $z$  sampled from prior distribution  $p_z$, the objective function can be expressed as:
\begin{equation}
\begin{split}
    \min_G \max_D V(D, G) &= \mathbb{E}_{x \sim p_{\text{data}}(x)}[\log D(x)] \\&+ \mathbb{E}_{z \sim p_z(z)}[\log(1 - D(G(z)))]
\end{split}
\end{equation}

Within industrial applications, GANs are widely used over the years \cite{74-mirza2021deep, 76-he2022survey}. This progression underscores the increasing versatility and significance of GANs in enhancing industrial processes through innovative AI-driven solutions.

\subsection{Diffusion Model}

Diffusion models, inspired by the physical diffusion process, have gained prominence in machine learning. The Denoising Diffusion Probabilistic Model (DDPM) \cite{106-ho2020denoising} has attracted significant academic attention, following the earlier score-based diffusion model proposed by Song et al. \cite{79-song2019generative}, later unified in SDE studies \cite{80-song2020score}. These models iteratively transform data into a simple distribution (e.g., Gaussian) and then reconstruct it via a two-stage Markov process.

For a given data distribution  $p(x)$ , the forward diffusion process perturbs data over time as:
\begin{equation}
    x_t = \sqrt{1 - \beta_t} x_{t-1} + \sqrt{\beta_t} \epsilon_t
\end{equation}
where  $\epsilon_t \sim \mathcal{N}(0, I)$  and  $\beta_t$  is a predefined noise factor. A denoising function  $q_\theta$  reconstructs  $x_{t-1}$  from  $x_t$ , leading to the DDPM loss function:
\begin{equation}
    \mathcal{L}(\theta) = \mathbb{E}_{x_t \sim p(x_t), \epsilon_t \sim \mathcal{N}(0, I)} \left[ -\log q_\theta(x_{t-1} | x_t) \right]
\end{equation}

Originally emerging in computer vision, diffusion models have expanded into sequential modeling \cite{83-cao2024survey}, audio processing \cite{84-lee2021nu}, and AI for scientific applications \cite{85-schneuing2022structure}. In industry, they have been applied to sensor data imputation \cite{88-chen2022dcpld} and anomaly detection \cite{6-xiao2023imputation}, addressing complex real-world challenges.

\section{The definition, characteristic, and demand of industrial time series generation}
In this section, we will present three points: 1) what is industrial time series generation 2) what are the characteristics of industrial time series generation 3) what are the current challenges of industrial time series acquisition, that evoke the demand for industrial generative models.

\subsection{The Definition of Industrial Time Series Generation}
The given time series \(T\) with \(N\) (\(N \geq 1\)) individual series of length \(L\) is represented as a matrix, i.e., \(T = (s_1, \ldots ,s_N)^T\), where each individual series \(s_i\) can be expressed as a \(L\)-dimensional vector, i.e., \(s_i = (x_{i,1}, \ldots , x_{i,L})\), and each \(x_{i,j}\) corresponds to a single time point \(t_j\) of \(s_i\). We denote \(p(s_1, \ldots ,s_N)\) as the real distribution of a given time series \(T\). In the context of industrial time series generation, \(L\)-dimensional vector variables represent $L$ industrial sensors. The goal of time series Generation is to create a synthetic time series \(T_{\text{gen}} = (s_{\text{gen},1}, \ldots ,s_{\text{gen},N})\) such that its distribution \(q(s_{\text{gen},1}, \ldots ,s_{\text{gen},N})\) is similar to \(p(s_1, \ldots ,s_N)\), and \(T_{\text{gen}}\) and \(T\) exhibit consistent statistical properties and patterns.

\subsection{The Characteristic of Industrial Time Series Generation}
Generative tasks in the industrial domain differ significantly from those in the internet domain, primarily in terms of data types and application scenarios. In the internet domain, generative tasks typically involve multimedia data such as text, images, and videos, with diverse applications like text-to-image and image-to-text generation. In contrast, industrial data is more diverse, including sensor data, text, and images, with the most critical being sensor time series data that records detailed operational states of equipment. Industrial time series generation has distinctive industrial characteristics.

1) Industrial time series have complex time series dependencies and patterns. As industrial processes become automated and industrial systems become more and more complex, it is no longer sufficient to rely on univariate time series data alone to provide a comprehensive and effective representation of industrial processes. As a result, multiple sensors are often utilized to monitor the entire industrial process. Generating multivariate time series is particularly challenging, as it requires dealing with correlations  and temporal dependencies between patterns as well as variables.

2) Dynamic variability inherent in industrial processes. Industrial processes typically exhibit a high degree of dynamic variability and are susceptible to conditions including pressure, temperature,  and humidity fluctuations. These environmental variations lead to fluctuations in outputs, resulting in data offsets and domain offsets. Therefore, to maintain accurate industrial data generation, these dynamic patterns must be captured.

3) Industrial scenarios require high reliability of time series. Sensor time series is the most common data type in the industry, so the subject of the generation task should be sensor time series data. In addition, industrial scenarios have strict requirements on the reliability of the data and the need to accurately present complex equipment operating states. Therefore, the generated time series data in the industrial scenarios must have authenticity and reliability to ensure accurate simulation of real industrial scenarios.

\subsection{Challenges in Industrial Time Series Acquisition}
There are a number of challenges in industrial time series acquisition that need to be addressed by generative models. 

\textbf{\textit{Limited Labeled Time Series}}: Collecting and testing data from complex industrial equipment such as aircraft engines is extremely difficult. This lead to the relatively small size of industrial well-labeled datasets. However, deep learning relies heavily on extensive labeled data for effective supervised learning, making it challenging to train models when labeled examples of industrial data are scarce. The amount of data can be augmented using generative models.

\textbf{\textit{Imbalanced Industrial Data Distribution}}: The problem of imbalanced industrial data distribution. In manufacturing systems, failures and anomalous events in industrial systems typically occur less frequently than in normal operating states, and there may be an insufficient number of instances of industrial equipment failures. This makes it difficult for deep learning models to accurately and efficiently capture and generalize patterns associated with these less representative states. Employing generative models enables the generation of a smaller number of samples, such as fault samples, resulting in a dataset with a balanced distribution.

\textbf{\textit{Missing Sensor Time Series Values}}: Missing data values are a common challenge in industrial time series applications. Unexpected interruptions in sensor failures and network delays can cause missing time series data. Managing and dealing with missing data is critical to maintaining the integrity of the time series model, as these missing values can interfere with the model learning process and affect the accuracy of the forecast. Generative models have the ability to fill in these missing values.

\textbf{\textit{Industrial Sensor Noise}}: Industrial environments are susceptible to sensor perturbations that introduce high-level noise into time series data. The time series disturbed by noise may not accurately reflect actual changes in equipment operation. Accurate differentiation between true features and noise is crucial for the modeling of deep learning models, hence specific strategies need to be implemented to mitigate the impact of sensor noise on time series applications in industrial. Generative models can generate noise-free sensor data through signal denoising methods.

\textbf{\textit{Industrial Privacy Concerns}}: Industrial data usually faces privacy protection issues. time series data collected by sensors and devices may contain private information or sensitive details about industrial processes. How to balance the need for data-driven insights with the imperative of protecting the privacy and security of industrial data is a great challenge in the industry. Generative models can address such issues by generating data that does not contain sensitive information.

\begin{figure*}[!htpb]
	\centering
	{
		\includegraphics[width=1.8\columnwidth]{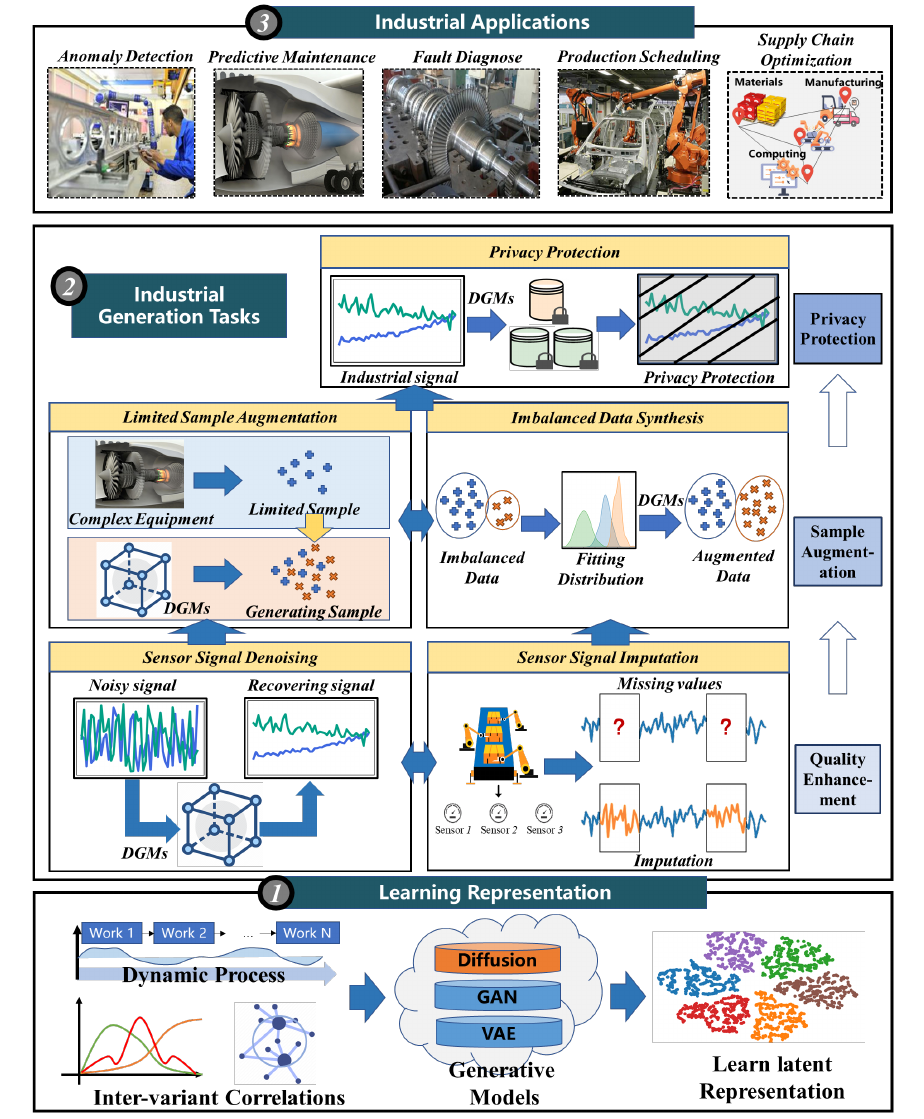}}
	\hfill    
	\caption{The DGM-baed AIGC framework of industrial time series generation. First, the generative model learns the latent representations of industrial time series by modeling dynamic processes and correlations between variables. Second, five specific generative tasks are executed in a downstream network. Finally, the generated samples will be applied to various industrial scenarios to address challenges such as sample scarcity in industrial scenarios.}
	\label{fig:Framework}
\end{figure*}

\section{The proposed DGM-based AIGC framework and the current state-of-the-art DGMs}

High-quality and adequate industrial time series are insufficient. In industrial systems, changes in equipment, environmental factors, and operating conditions, can lead to fluctuations in monitoring data. This variability and complexity create challenges for the data collection of complex equipment, leading to a lack of labeled datasets and an imbalanced data distribution. In addition, manufacturing systems may experience unexpected interruptions in sensor failures, network delays, and data transmission, leading to problems of high-level noise and missing values. In decentralized industrial collaborations such as cloud manufacturing, industrial data from different users remain difficult to share due to privacy concerns. These challenges make it difficult for industrial data to meet the demands of deep learning methods for time series applications in the industry.

The DGM-baed AIGC framework is proposed for industrial data generation, as illustrated in Fig. \ref{fig:Framework}. First, generative models learns the latent representations of industrial time series by modeling dynamic processes and correlations between variables, which have been introduced in Section II and Section III. Second, quality enhancement, sample augmentation, and privacy protection are performed to improve the quality and quantity of data, which will be discussed in Section IV. Finally, the generated samples will be applied to various industrial scenarios to address challenges such as sample scarcity in industrial scenarios, as presented in Section V. With the development of generative intelligence, there are many methods are proposed to tackle these issues. 

\subsection{Limited Sample Augmentation} 

Limited sample augmentation utilizes generative methods to enhance datasets where sample sizes are insufficient for effective model training. This approach is critical in scenarios where acquiring real-world data is challenging, enabling the creation of realistic and diverse samples to improve model accuracy and generalization.

In industrial applications, this technique has significantly improved fault diagnosis and anomaly detection. A bidirectional alignment network with VAE addresses data scarcity in thermal power plant fault diagnosis, achieving generalization across rare fault categories \cite{20-li2023federated}. MSGAN combines offline training with real-time inference, using an adaptive update strategy and gradient penalty to generate high-quality anomaly samples, enhancing anomaly detection in robotic sensors \cite{55-lu2021gan}. Recently, optimized diffusion models leveraging asymptotic denoising and temporal dynamics extraction \cite{GUO2025115951, ren2024iiot, Ren2024diffmts} have outperformed GANs in predictive maintenance, further demonstrating their versatility in limited sample augmentation.

\subsection{Imbalanced Data Synthesis} 

Imbalanced data synthesis involves generating artificial data to address class imbalances in datasets, a crucial technique in machine learning and industrial applications, especially for underrepresented categories.

In industry, several innovative approaches tackle data imbalance. Fan et al. \cite{2-fan2023effective} proposed a VAE-based method for semiconductor fault detection, effectively mitigating the scarcity of defective wafer samples. DRL-GAN integrates reinforcement learning with GANs, achieving balanced data distributions \cite{4-benaddi2022anomaly}. Recently, diffusion models have gained traction in fault diagnosis. Yan et al. \cite{46-yan2023bearing} introduced a noise-to-signal inversion mechanism for bearing fault detection, preserving spectral characteristics. PSDM enhances feature consistency in sparse fault patterns through adaptive weight sharing \cite{xiao2024parameter}. Additionally, digital twin architectures combined with diffusion processes embed physical constraints via multi-physics simulations, improving rotating machinery diagnostics \cite{jiang2024novel}.

\subsection{Sensor Signal Imputation} Sensor signal imputation describes the process of estimating and filling in missing or incomplete data in datasets collected from sensors. The accuracy and completeness of sensor data are essential for reliable analysis and decision-making.

Several advanced methods have been developed for sensor data imputation. Velasco-Gallego et al. \cite{11-velasco2022analysis} introduced a VAE-based method for marine machinery sensor data, while FedTMI \cite{12-yao2022fedtmi} applies edge-computed GANs and federated transfer learning to industrial missing data. Among GAN-based approaches, models like IM-GAN \cite{13-wu2022imputing} and SGAIN \cite{14-gao2022missing} tackle complex multivariate time series and structural health monitoring, while Semi-GAN \cite{16-lee2022semi} and ST-GAIN \cite{17-ren2022spatio} improve imputation in semiconductor and manufacturing data. DiffAD \cite{6-xiao2023imputation} integrates denoising diffusion with a multi-scale state space model for time series anomaly detection, capturing long-term dependencies. GD-GRU \cite{19-Xu2021DeepLearning} combines a Gaussian diffusion process with GRU, enhancing environmental data repair.

\subsection{Sensor Signal Denoising} To enhance the quality and reliability of the information captured by sensors, sensor signal denoising is used to remove noise from sensor data. The primary objective is to filter out these extraneous noises without distorting the actual signal. 

Sensor Signal Denoising has been explored through various models, including AR, GAN, and hybrid approaches. AR models have demonstrated effectiveness in denoising, with Liu et al. \cite{24-Liu2023HarmonicReducerb} integrating wavelet regional correlation thresholding and ARMA models to enhance acoustic emission signal clarity in industrial robots. In GAN-based methods, CycleGAN has been applied with Bayesian nonparametric estimation for online sensor signal denoising in milling processes \cite{27-Wei2023MultisensorSignalsa}. Additionally, Att-DCDN leverages an encoder-decoder GAN structure to integrate synthetic and field seismic data, improving seismic noise attenuation. Min et al. \cite{54-min2021ddae} further enhance seismic data denoising by combining a pre-trained deep denoising autoencoder with transfer learning, optimizing training on limited field data.

\begin{table*}[!htpb]
\centering
\caption{Multi-perspective categorization of industrial time series DGMs from three criteria: industrial generation tasks, applications, and model architectures. The following abbreviations are used in the architecture column: AR (Autoregressive model), VAE (Variational Autoencoder), GAN (Generative Adversarial Network).}
\setlength{\tabcolsep}{4.6mm}{
\begin{tabular}{lllcc}
\hline
Industrial Generation Task                                          & Paper                 & \multicolumn{1}{c}{Year} & Application                                                                       & Architecture     \\ \hline
\multirow{20}{*}{Limited Sample Augmentation} & Moreno-Barea et al.\cite{29-Mor2020ImprovingClassificationa} & 2020                     & General Industrial Applications                                                   & VAE+GAN          \\
& MSGAN\cite{55-lu2021gan}              & 2021                     & Anomaly Detection                                                                 & GAN              \\
& Hoh et al.\cite{5-hoh2022generative}           & 2022                     & Anomaly Detection                                                                 & GAN              \\
& DCGAN\cite{7-gao2022deep}               & 2022                     & General Industrial Applications                                                   & GAN              \\
& MAS-GAN\cite{8-zhang2022novel}             & 2022                     & General Industrial Applications                                                   & GAN              \\
& Chen et al.\cite{15-chen2022repairing}         & 2022                     & General Industrial Applications                                                   & GAN              \\
& Upadhyaya et al.\cite{59-upadhyaya2022demand}   & 2022                     & \begin{tabular}[c]{@{}c@{}}Supply Chain and\\      Production Scheduling\end{tabular} & GAN              \\
& DA-JITL\cite{30-Jia2022ImprovingPerformancea}            & 2022                     & General Industrial Applications                                                   & VAE              \\
& Huang et al.\cite{9-huang2023hybrid}         & 2023                     & Anomaly Detection                                                                 & Diffusion Models \\
& Li et al.\cite{20-li2023federated}           & 2023                     & Fault Diagnosis                                                                   & VAE              \\
& Souza et al.\cite{28-Bap2023DataAugmentationa}        & 2023                     & Fault Diagnosis                                                                   & AR Models        \\
& CPI-GAN\cite{47-xiong2023controlled}            & 2023                     & Predictive Maintenance                                                            & GAN              \\
& Seq2Seq-WGAN\cite{61-zhang2023improving}       & 2023                     & \begin{tabular}[c]{@{}c@{}}Supply Chain and\\      Production Scheduling\end{tabular} & GAN              \\
& Liu et al.\cite{60-liu2023generative}          & 2023                     & \begin{tabular}[c]{@{}c@{}}Supply Chain and\\      Production Scheduling\end{tabular} & GAN      \\

& TimeDDPM\cite{52-dai2023timeddpm}           & 2023                     & General Industrial Applications  & Diffusion Models \\ 
& ATT-TSAAE.\cite{chen2024two} & 2024 & Anomaly Detection  & GAN+Transformer \\ 
& Time-Transformer\cite{liu2024time}  & 2024 & General Industrial Applications  & Transformer \\ 
& DiffUCD\cite{GUO2025115951} & 2025 & Fault Diagnosis  & Diffusion Models \\  
& SDE Diffusion \cite{ren2024iiot} & 2024 & Predictive Maintenance   & Diffusion Models \\
& Diff-MTS\cite{Ren2024diffmts} & 2024 & Predictive Maintenance   & Diffusion Models \\  

\hline
\multirow{18}{*}{Imbalanced Data Synthesis}   & WGAN-SAE\cite{32-Wan2019GeneralizationDeepa}           & 2019                     & Fault Diagnosis                                                                   & GAN              \\
  & Ortego et al.\cite{33-Ort2020DataAugmentationb}       & 2020                     & Predictive Maintenance                                                            & GAN              \\
  & Gao et al.\cite{34-Gao2020DataAugmentationb}          & 2020                     & Fault Diagnosis                                                                   & GAN              \\
  & GAN-GDA\cite{36-Zhu2020GaussianDiscriminativeb}            & 2020                     & Fault Diagnosis                                                                   & GAN              \\
  & Oh et al.\cite{45-oh2020imbalanced}           & 2020                     & Predictive Maintenance                                                            & GAN              \\
  & AWL-VAE\cite{3-nakata2022proposal}             & 2022                     & Anomaly Detection                                                                 & VAE              \\
  & DRL-GAN\cite{4-benaddi2022anomaly}             & 2022                     & Anomaly Detection                                                                 & GAN              \\
  & NCVAE-AFL\cite{21-Zha2022NormalizedConditionala}          & 2022                     & Fault Diagnosis                                                                   & VAE              \\
  & WCGAN-HFM\cite{22-Pen2022NovelBearing}          & 2022                     & General Industrial Applications                                                   & GAN              \\
  & Liu et al.\cite{31-Liu2022IntrusionDetectiona}          & 2022                     & General Industrial Applications                                                   & VAE              \\
  & Fan et al.\cite{2-fan2023effective}           & 2023                     & Anomaly Detection                                                                 & VAE              \\
  & Song et al.\cite{23-Son2023DeepGenerative}         & 2023                     & General Industrial Applications                                                   & VAE + GAN        \\
  & DB-CGAN\cite{35-Zho2023DistributionBiasb}           & 2023                     & General Industrial Applications                                                   & GAN              \\
  & Yan et  al.\cite{46-yan2023bearing}         & 2023                     & Fault Diagnosis                                                                   & Diffusion Models \\ 
  & Yang et al.\cite{Yang2024DIFF} & 2024 & Fault Diagnosis  & Diffusion Models \\ 
  & PSDM\cite{xiao2024parameter} & 2024                     & Fault Diagnosis     & Diffusion Models \\ 
  & Jiang et al.\cite{jiang2024novel} & 2024                     & Fault Diagnosis     & Diffusion Models \\ 
  & MBAC-GAN\cite{Yu2025GAN} & 2025                     & Fault Diagnosis    & GAN \\ \hline
\multirow{11}{*}{Sensor Signal Imputation}      & GD-GRU\cite{19-Xu2021DeepLearning}             & 2021                     & General Industrial Applications                                                   & Diffusion Models \\
      & Sarda et al.\cite{48-sarda2021missing}        & 2021                     & General Industrial Applications                                                   & GAN              \\
      & Velasco et al.\cite{11-velasco2022analysis}      & 2022                     & Predictive Maintenance                                                            & VAE              \\
      & FedTMI\cite{12-yao2022fedtmi}             & 2022                     & General Industrial Applications                                                   & GAN              \\
      & IM-GAN\cite{13-wu2022imputing}             & 2022                     & General Industrial Applications                                                   & GAN              \\
      & SGAIN\cite{14-gao2022missing}              & 2022                     & Predictive Maintenance                                                            & GAN              \\
      & Semi-GAN\cite{16-lee2022semi}           & 2022                     & General Industrial Applications                                                   & GAN              \\
      & ST-GAIN\cite{17-ren2022spatio}            & 2022                     & General Industrial Applications                                                   & GAN              \\
      & FIGAN et al.\cite{49-yao2021figan}        & 2022                     & General Industrial Applications                                                   & GAN              \\
      & DiffAD\cite{6-xiao2023imputation}              & 2023                     & Anomaly Detection                                                                 & Diffusion Models \\
      & SGT-GAIN.\cite{18-Li2023TransformerenabledGenerative}           & 2023                     & General Industrial Applications                                                   & GAN              \\ \hline

\multirow{6}{*}{Sensor Signal Denoising}      
      & Att-DCDN.\cite{51-wang2020attribute}           & 2021                     & General Industrial Applications                                                   & GAN              \\
      & DDAE-GAN.\cite{54-min2021ddae}           & 2021                     & General Industrial Applications                                                   & GAN              \\
      & AE-NAR.\cite{1-zhang2022anomaly}             & 2022                     & Anomaly Detection                                                                 & AR + AE          \\
      & Liu et al.\cite{24-Liu2023HarmonicReducerb}          & 2023                     & General Industrial Applications                                                   & AR               \\
      & Su et al.\cite{25-Su2024ForecastingNatural}          & 2023                     & General Industrial Applications                                                   & AR               \\
      & UKF-CycleGAN\cite{27-Wei2023MultisensorSignalsa}       & 2023                     & General Industrial Applications                                                   & GAN              \\ \hline
\multirow{11}{*}{Privacy Protection}          & Bernieri et al.\cite{40-Ber2019KingFisherIndustriala}     & 2019                     & Anomaly Detection                                                                 & VAE              \\
      & Keshk et al.\cite{38-Kes2020PrivacyPreservingFrameworkBasedBlockchain}        & 2020                     & Anomaly Detection                                                                 & VAE              \\
      & Augenstein et al.\cite{43-augenstein2019generative}   & 2020                     & General Industrial Applications                                                   & GAN              \\
      & Palekar et al.\cite{37-Pal2022APROSecreta}      & 2022                     & General Industrial Applications                                                   & AR               \\
      & Almaiah et al.\cite{39-Alm2022LightweightHybrida}     & 2022                     & General Industrial Applications                                                   & VAE              \\
      & Fed-LSGAN\cite{42-li2021privacy}          & 2022                     & General Industrial Applications                                                   & GAN              \\
      & ProcessGAN\cite{57-li2022generating}         & 2022                     & General Industrial Applications                                                   & GAN              \\
      & Chen et al.\cite{41-chen2021private}         & 2023                     & General Industrial Applications                                                   & GAN              \\
      & SDIN\cite{50-wu2023deep}               & 2023                     & General Industrial Applications                                                   & GAN              \\
      & DPDM\cite{53-chu2023differentially}  & 2023         & General Industrial Applications  & Diffusion Models \\
      & Wang et al.\cite{56-wang2022federated}   & 2023    & General Industrial Applications   & VAE              \\ \hline
\end{tabular}}
\end{table*}

\subsection{Privacy Protection} In the context of the industrial sector and generative models, privacy protection refers to safeguarding sensitive and proprietary information during the deployment and application of these advanced computational models.

Palekar et al. \cite{37-Pal2022APROSecreta} introduced an AR-based authentication model (APRO), combining secret key generation and Box-Cox transformation for secure, efficient encryption in industrial devices. The KingFisher framework \cite{40-Ber2019KingFisherIndustriala} utilizes VAEs to analyze network interactions and enhance privacy, while a federated learning mechanism \cite{56-wang2022federated} applies VAEs with differential privacy noise for secure medical IoT data. In GAN-based methods, Chen et al. \cite{41-chen2021private} proposed a differentially private GAN to protect machine operation data, while Fed-LSGAN \cite{42-li2021privacy} integrates federated learning with LSGANs for privacy-aware energy scenario generation. Recently, a differentially private DDPM \cite{53-chu2023differentially} has been developed, incorporating differential privacy into diffusion models to generate high-quality synthetic data while preserving privacy.

\section{Typical Industrial Applications of DGMs}
After discussing the main tasks of DGMs in industry, in this section, we mainly introduce the applications in industrial scenarios.

\subsection{Anomaly Detection} 
In industry, anomaly detection identifies irregular patterns in data, signaling equipment failures, inefficiencies, or security risks. It is essential for efficiency, safety, and minimizing downtime, using machine learning and deep learning to analyze sensor data, network traffic, and production metrics.
The AE-NAR model enhances wind turbine pitch-bearing anomaly detection by combining an autoencoder for global feature extraction with nonlinear autoregression for noise reduction \cite{1-zhang2022anomaly}. The adaptive weighted loss VAE improves accuracy in industrial anomaly detection by dynamically adjusting loss weights \cite{3-nakata2022proposal}. Similarly, conditional GANs process high-frequency time-series data in smart manufacturing \cite{5-hoh2022generative}.
Additionally, Chen et al. propose a two-stage adversarial Transformer-based method, integrating an autoregressive TCN and adversarial training, achieving superior performance on real-world industrial datasets \cite{chen2024two}. Alongside other GANs and Diffusion models \cite{4-benaddi2022anomaly, 6-xiao2023imputation, 55-lu2021gan}, these approaches highlight the growing role of complex computational models in industrial anomaly detection.

\subsection{Predictive Maintenance} Predictive maintenance is a strategic approach that employs data analytics, machine learning, and advanced computational models to predict and prevent equipment failures before they occur. This approach aims to optimize maintenance operations by identifying potential issues early, thereby reducing unplanned downtime and extending equipment life. 
A GAN-based data augmentation method addresses fault data scarcity in Industry 4.0, improving machine learning models for failure prediction \cite{33-Ort2020DataAugmentationb}. Expanding on this, Gaussian process regression with GAN enhances missing data handling for quality assurance and maintenance scheduling \cite{45-oh2020imbalanced}.
For RUL prediction, CPI-GAN synthesizes degradation trajectories, integrating physics-informed constraints for accurate maintenance scheduling \cite{47-xiong2023controlled}. Additionally, VAEs contribute to predictive maintenance by diversifying computational strategies \cite{11-velasco2022analysis}.
Furthermore, Diffusion models improve predictive maintenance by enabling stable industrial time series generation. SDE diffusion with PC samplers enhances reliability \cite{ren2024iiot}, while Diff-MTS integrates Ada-MMD and TDR-UNet for superior temporal pattern learning, outperforming GAN-based methods \cite{Ren2024diffmts}.

\subsection{Fault Diagnosis} Fault diagnosis refers to the process of identifying, analyzing, and rectifying faults in industrial equipment and systems. It leverages advanced data analytics, machine learning algorithms, and sensor data to accurately and promptly diagnose equipment failures. 
Zhao et al. propose NCVAE-AFL to address class imbalance in bearing-rotor fault diagnosis \cite{21-Zha2022NormalizedConditionala}. A time-varying autoregressive model enhances fault detection in non-stationary multivariate time series \cite{28-Bap2023DataAugmentationa}.
WGAN-based approaches improve fault diagnosis accuracy by tackling mechanical fault data scarcity \cite{32-Wan2019GeneralizationDeepa,34-Gao2020DataAugmentationb}, while GAN-GDA integrates Gaussian discriminant analysis for industrial fault detection \cite{36-Zhu2020GaussianDiscriminativeb}. 
Furthermore, Diffusion models enhance fault diagnosis by generating unknown condition fault signals (DiffUCD \cite{GUO2025115951}), improving feature diversity (PSDM \cite{xiao2024parameter}), and creating realistic fault data for digital twin diagnostics \cite{jiang2024novel}. A diffusion-based augmentation method ensures stable sample generation \cite{Yang2024DIFF}.
These studies emphasize the key role of machine learning and data analytics in industrial fault detection.

\subsection{Supply Chain Optimization and Production Scheduling} Supply chain optimization and production scheduling are key concepts in operations management, focused on enhancing the efficiency and effectiveness of the entire supply chain through rational resource allocation, coordination of production activities, and process improvement. 
Recent advances leverage GANs and VAEs for deep-learning-based industrial optimization. CVAE aids automotive manufacturing by generating health indicators for machine degradation and efficiency \cite{58-zhai2021enabling}. Deep generative models optimize supply chain management by generating creative alternatives \cite{59-upadhyaya2022demand}.
A TimeGAN-CNN-LSTM hybrid model improves short-term load forecasting for energy management \cite{60-liu2023generative}, while Seq2Seq-WGAN enhances cement production scheduling by capturing temporal dependencies and generating f-CaO label data \cite{61-zhang2023improving}.

\begin{figure*}[!htpb]
	\centering
	{
		\includegraphics[width=2\columnwidth]{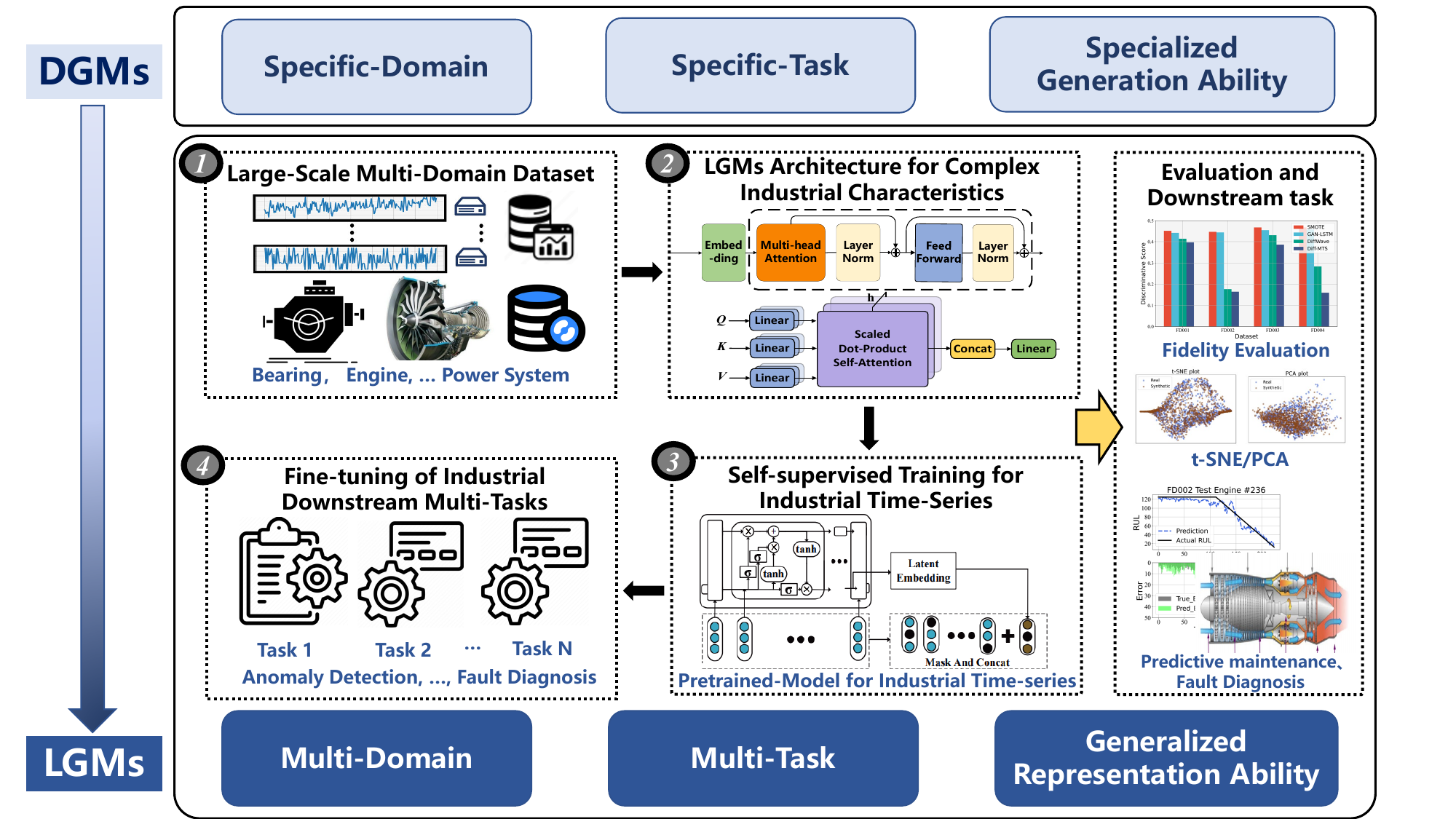}}
	\hfill    
	\caption{From DGMs to LGMs, the research methods of generative models have undergone profound changes. The roadmap for building large generative models from four aspects: Large industrial dataset, LGMs architecture for complex industrial characteristics, self-supervised training for industrial time series, fine-tuning for industrial downstream tasks.}
	\label{fig:LGMs}
\end{figure*}

\section{Large Generative Models for Industrial Time Series}
In the industrial domain, generative models have made significant progress but their usefulness is limited. In complex and open industrial environments, generative models may face several challenges, including limited generalization capabilities, multi-tasking limitations, and cognitive limitations. Specifically, although generative models perform well in known scenarios, they struggle with the complexity of real industrial scenarios due to insufficient generalization capabilities. In addition, current generative models are usually designed for a single task. While industrial equipment usually contains hundreds of core components, developing a corresponding generative model for each core component separately is difficult to achieve. Finally, generative models have limited cognitive ability to understand the nature of industrial data, which causes them to potentially generate incomprehensible and erroneous results.

The generative AI research is shifting from the traditional paradigm of specific-domain, specific-task, and specialized generation ability to the new paradigm of multi-domain, multi-task, and generalized generation ability\cite{ren2025sciifm,Ren2025IFM}, such as ChatGPT\cite{109-ouyang2022training} and DALL·E 2\cite{110-ramesh2022hierarchical}. However, there is still no definitive answer on how to construct the large generative model in industrial time series. To promote the research and application of large generative models in industry, we explores theoretically how to build large generative models suitable for industrial applications, as illustrated in Fig. \ref{fig:LGMs}.

\subsection{Large-Scale Industrial Dataset}
Large-scale datasets are the cornerstone to drive large generative model research. Unlike the NLP or the CV, the industrial time series are generally sensor readings of time series data, such as pressure, temperature, speed, vibration signals, and power signals. 
There are a number of open-source industrial datasets of various sizes and domains, such as bearing failure dataset\cite{wang2018hybrid}, wind turbine dataset\cite{zappala2019electrical}, three-phase motor failure dataset\cite{jung2023vibration}, turbofan engine degradation dataset\cite{arias2021aircraft}, gearbox failure dataset\cite{shao2018highly}, etc.  Specifically, the CWRU\cite{smith2015rolling} dataset is a classical fault diagnosis dataset that records bearing vibration data under different operating conditions, including parameters such as different rotational speeds, loads, and operating times. It consists of four different failure modes, including inner ring failure, outer ring failure, and rolling element failure. The CMAPSS dataset\cite{saxena2008damage} records the turbofan engine degradation monitoring data, which contains six operating conditions and 2 failure states. 

Although LGMs have accomplished impressive results in the NLP and CV, industrial data involves various types of multi-source devices and sensors with complex device information. Therefore, self-supervised representation learning methods need to be designed according to the unique characteristics of industrial data in order to make the LGMs applicable in this field. For example, an adaptive sensor weighting method\cite{ren2023time} based on time-varying Gaussian encoders has been proposed to construct representative features of industrial time series. The article\cite{ren2021data} proposes an LSTM-DeepFM model for solving industrial soft sensor problems, which utilizes an LSTM-autoencoder in the pre-training phase to enhance the feature construction process.

\subsection{LGMs Architecture Tailored for Industrial Characteristics}
Currently, the most representative LGMs include GPT-4\cite{114-achiam2023gpt} and DALL·E 2\cite{110-ramesh2022hierarchical}, of which the linguistic generative model GPT-4 adopts Transformer as its core architecture, while the multimodal generative model DALL·E uses the Diffusion Model. Diffusion model is powerful in handling probability distributions and generating samples. Transformer excels in powerful sequence modeling capabilities, and its self-attention mechanism allows it to process time series data in parallel while efficiently capturing important feature information. Transformer has also made notable achievements in industrial time series\cite{ren2023dlformer,xu2022zero}. However, one of the keys to deal with industrial time series data is efficient modeling of complex systems and extracting time series dependencies. For LGMs architecture, future research could focus on optimizing attention mechanisms, and decomposing and transforming time series data to improve representation and modeling capabilities.

\textbf{Handling Dynamic Dependencies with Adaptive Attention Mechanisms:} Adaptive attention mechanisms aim to refine how attention is distributed over time-series data, focusing on the most relevant time steps. These mechanisms adjust the attention weights based on the dynamic nature of industrial data, helping capture both temporal dependencies\cite{Ren2024diffmts} (e.g., recent trends) and spatial dependencies\cite{ren2023time} (e.g., multi-sensor correlations) effectively. 

\textbf{Domain-Specific Embeddings:} Industrial time-series data requires specialized embedding techniques to better represent features like sensor readings over time. Position embeddings can be adapted to handle irregular time intervals, while patch embeddings\cite{Nie2023PatchTST} can be used to segment time-series data into meaningful subsequences, improving the model's ability to extract local temporal patterns. 

\textbf{Frequency Domain Transformations:} Frequency domain transformations such as Fast Fourier Transform (FFT) and wavelet transforms\cite{peng2004application} can be applied to industrial time-series data to highlight periodic patterns or anomalies not easily detectable in the raw time domain. These transformations enable the model to focus on important frequency components, improving predictive accuracy.

\color{black}
\subsection{Self-supervised Training of Industrial Time Series}
Self-supervised learning is an approach to unsupervised learning that allows a model to learn its own representation from input data without external labels. As mentioned earlier, when building industrial LGMs, a large amount of data is usually required to train the model. However, not all of the data is well-labeled. Self-supervised learning enables models to extract valuable intrinsic features from unlabeled data, thus making more efficient use of large-scale datasets. This approach has become one of the core algorithms for building large models. Currently, there are some self-supervised learning methods that have been proven successful in processing industrial data. The article\cite{ren2021data} proposes an LSTM-DeepFM model, which utilizes LSTM-autoencoder for self-supervised learning to enhance the feature construction process.

The industrial environment involves a variety of heterogeneous devices that may provide different types of data, including temperature, pressure, flow rate, etc. One challenge of self-supervised learning is to process heterogeneous device information and integrate it into a consistent representation. This enables the model to understand data from different devices and achieve pattern generalization. Therefore, domain adaptation and transfer learning approaches can be employed to process heterogeneous information from multiple sources in order to learn domain-independent generalized feature representations. For example, Ren et al.\cite{ren2023single} proposed cross-domain learning methods of multi-source black-box data. It achieved cross-domain common mechanism learning in privacy and security scenarios without touching the source domain task data. 

\subsection{Fine-tuning for Industrial Downstream Tasks}
As described in Sections III and IV, industrial generative models are usually applied to a variety of tasks, including time series imputation, generation, denoising, and so on. In addition, generative models are used in a variety of application areas, including fault diagnosis, health management, and anomaly detection. At present, most of the generative models target a single task or a single domain. In multi-task and multi-domain scenarios, designing generative models for each task or application domain and training them separately requires a lot of resources. Some approaches have explored the multi-task learning to enhance the model performance. Liu et al.\cite{liu2020multitask} proposed a multi-task one-dimensional convolutional neural network, which combines the primary bearing fault diagnosis task (FDT) and auxiliary tasks to realize shared feature learning, thus improving the bearing fault diagnosis performance.

Industrial generative models still need to make efforts in multi-task adaptation, and the main fine-tuning methods for large models are as follows:
(1) Task-oriented model fine-tuning: this approach typically builds on a pre-trained model and uses task-specific data for further training. This can be achieved by tuning the model parameters, adding new layers, or modifying the loss function. Through this fine-tuning, the model will learn task-specific fine-grained features and representations.
(2) Model fine-tuning with prompt learning: Designing templates that fit the upstream pre-training task taps the potential of the pre-trained model, allowing the model to perform the downstream task better with as little labeled data as possible. 
(3) Model fine-tuning based on adapter networks: adapter networks are lightweight network structures that can be inserted into pre-trained models for learning task-specific representations. Adapters can add additional task-related information to the model without destroying the knowledge of the pre-trained model.

\section{Evaluation Metrics}  
The quality of time series generation models is typically evaluated based on four core principles: fidelity, diversity, utility, and computational efficiency. This section systematically reviews relevant evaluation metrics and categorizes their principles and significance.

\subsection{Fidelity Evaluation}  
Fidelity measures the consistency between generated and original time series in terms of statistical properties, dynamic patterns, and local context.  
  \subsubsection{Model-based Methods}  
  Indirectly assess generation quality through downstream models:

\begin{itemize}  
\item \textbf{Discriminative Score (DS)\cite{yoon2019time}}: A 2-layer LSTM classifier is trained to distinguish between real and generated time series. The classification error reflects the realism of the generated time series.  
\item \textbf{Predictive Score (PS)\cite{yoon2019time}}: A time series prediction model (e.g., GRU/LSTM) is trained on generated time series and tested on original data using metrics like MAE, evaluating the preservation of temporal dynamics.  
\item \textbf{Contextual-FID (C-FID)\cite{jeha2022psa}}: Extends the FID metric from image generation to time series by computing the Fréchet distance between feature distributions of generated and real time series using context-aware embeddings.  
\end{itemize}  

  \subsubsection{Feature-based Methods}  
  Directly compare statistical feature differences. 
  \begin{itemize}  
    \item \textbf{Marginal Distribution Difference (MDD\cite{ni2021sig})}: Computes the average absolute difference between histograms of original and generated time series, assessing distribution alignment.  
    \item \textbf{Correlation Score (CS)\cite{liao2020conditional}}: Measures the difference in correlation coefficients between original and generated time series, evaluating the preservation of temporal dependencies.   
  \end{itemize}  

  \subsubsection{Distance-based Methods}  
  Provide deterministic metrics to avoid model randomness.  
  \begin{itemize}  
    \item \textbf{Euclidean Distance (ED)\cite{ang2023tsgbench}}: Computes the mean pointwise Euclidean distance between generated and original time series, suitable for normalized data.  
    \item \textbf{Dynamic Time Warping (DTW)\cite{Ren2024diffmts}}: Captures similarity in asynchronous temporal patterns by finding the optimal alignment path, robust to temporal shifts.  
  \end{itemize}  

\subsection{Diversity Evaluation}  
Diversity ensures that generated time series cover the original data distribution and avoid mode collapse:  
\begin{itemize}  
  \item \textbf{t-SNE/PCA Visualization}: Projects high-dimensional time series into 2D space to visually compare the distribution coverage and cluster structures of generated and real data points.  
  \item \textbf{Distribution Plot Analysis}: Plots kernel density estimates of key statistics (e.g., mean, variance, extremes) to verify the alignment between generated and original data distributions.  
\end{itemize}  

\subsection{Utility Evaluation}  
Assesses the practical utility of generated data in real-world tasks:  
\begin{itemize}  
  \item \textbf{Predictive Performance (PS)\cite{yoon2019time}}: As described in fidelity evaluation, PS also reflects the ability of generated data to support downstream prediction tasks.  
  \item \textbf{Downstream Task Validation}: We consider adopting the Train on Synthetic and Real, Test on Real (TSRTR) paradigm to train classification/regression models on generated data and evaluate their performance on real test sets, directly validating the practical value of generated data.  
\end{itemize}  

\subsection{Computational Efficiency Evaluation}  
\begin{itemize}  
  \item \textbf{Training Time}: Records the wall-clock time from model initialization to convergence, particularly focusing on scalability for large-scale data.  

\end{itemize}

\subsection{Analysis}
Model-based methods, such as DS and PS, are the most commonly used for evaluating the fidelity of generated time series. However, these methods are sensitive to model architecture, hyperparameters, and training randomness, which can lead to inconsistent results. To mitigate these issues, multiple runs with result averaging are recommended. While visualization methods like t-SNE/PCA offer valuable insights, they rely on subjective interpretation and may not provide precise quantitative assessments. Comprehensive evaluation should be implemented to provide a more reliable and deterministic evaluation. Ultimately, the gold standard for evaluating industrial time series generation is the effectiveness of the generated data in downstream tasks, assessing its practical performance.

\section{Case Study: Application of DGMs in Engine Predictive Maintenance}
In this case study, we implement a type of deep generative model, the diffusion model, to enhance downstream tasks in aircraft engine maintenance. Specifically, we focused on improving predictive models by generating synthetic data and combining it with real-world sensor data. 

\begin{figure}[!htpb]
	\centering
	{
		\includegraphics[width=1\columnwidth]{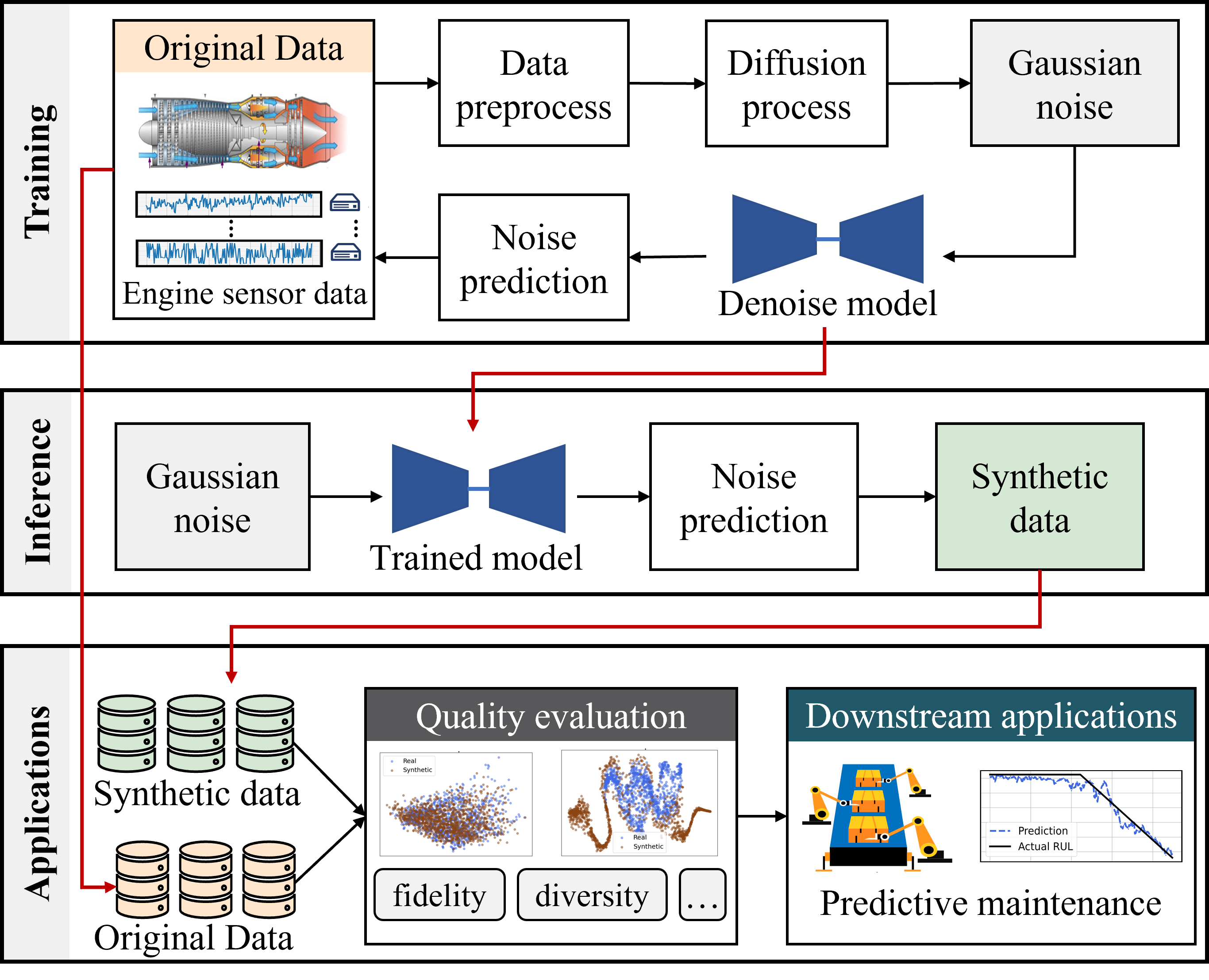}}
	\hfill    
	\caption{Workflow of implementing diffusion model in this case study}
	\label{fig:Workflow}
\end{figure}

\subsection{Workflow}
As illustrated in Fig. \ref{fig:Workflow}, the workflow outlines the steps for deploying a diffusion model-based approach to generate synthetic industrial data for a case study focused on predictive maintenance of aircraft engine sensors.

\textbf{Step 1: Learning representation of industrial time series in the training phase}
In the training phase, the process begins with the collection of original engine sensor data, which undergoes initial preprocessing to prepare it for further analysis. This preprocessing step typically involves normalization and cleaning, ensuring that the data is in a suitable format for the model. Once the data is preprocessed, a diffusion process is applied. In this stage, Gaussian noise is progressively added to the original data across multiple time steps, simulating the degradation of the data. This process is mathematically described by:
\[ q(x_t | x_{t-1}) = \mathcal{N}(x_t; \sqrt{1 - \beta_t} x_{t-1}, \beta_t I) \]
where the noise \(\beta_t\) is introduced at each time step to corrupt the original data. Following the noise addition, the model is trained to predict and reverse this noise. A \textbf{denoising model} is employed, typically utilizing a U-Net architecture to learn how to recover the original data by progressively removing the noise added during the diffusion process. The model is trained to minimize the difference between the predicted noise and the actual noise added, refining its ability to reconstruct the data accurately.

\textbf{Step 2: Limited sample augmentation  in the inference phase}
During inference, the trained model starts with random Gaussian noise. Through iterative denoising, the model generates synthetic data that mimics the original engine sensor data. The denoising process is mathematically represented by:

\[
x_{t-1} = \mu_{\theta}(x_t, t) + \sigma_t z, \quad z \sim \mathcal{N}(0, I)
\]

where the model predicts the noise at each step and adjusts the data accordingly.

\textbf{Step 3: Evaluation and industrial application}
Once the synthetic data is generated, it undergoes quality evaluation to ensure it aligns well with the statistical properties of the original data. Metrics such as fidelity and diversityare used to assess the quality of the synthetic data, where fidelity measures how closely the synthetic data matches the original, and diversity reflects the range of variations within the generated dataset. The synthetic data generated through this process is then used for downstream applications. In this case study, the generated data is specifically applied to aircraft engine predictive maintenance.  By using synthetic data, predictive models can be evaluated and improved, ultimately aiding in the early detection of potential engine failures, optimizing maintenance schedules, and improving overall reliability in aircraft engine operations.

\subsection{Experiment Setup}
We conducted experiments using the CMAPSS engine datasets, which include four subsets: FD001, FD002, FD003, and FD004. These datasets contain time-series sensor data from aircraft engine systems. We deployed the Diff-MTS model\cite{Ren2024diffmts} and evaluated its performance against other DGMs and the traditional Synthetic Minority Over-sampling Technique (SMOTE\cite{chawla2002smote}). The evaluation focused on fidelity, diversity, and usability of the synthetic data. Finally, we implemented an industrial downstream task focused on predictive maintenance, evaluating the impact of augmented data on predictive accuracy.

\subsection{Experiment Results}
\begin{figure}[!htpb]
	\centering
	{
		\includegraphics[width=1.12\columnwidth]{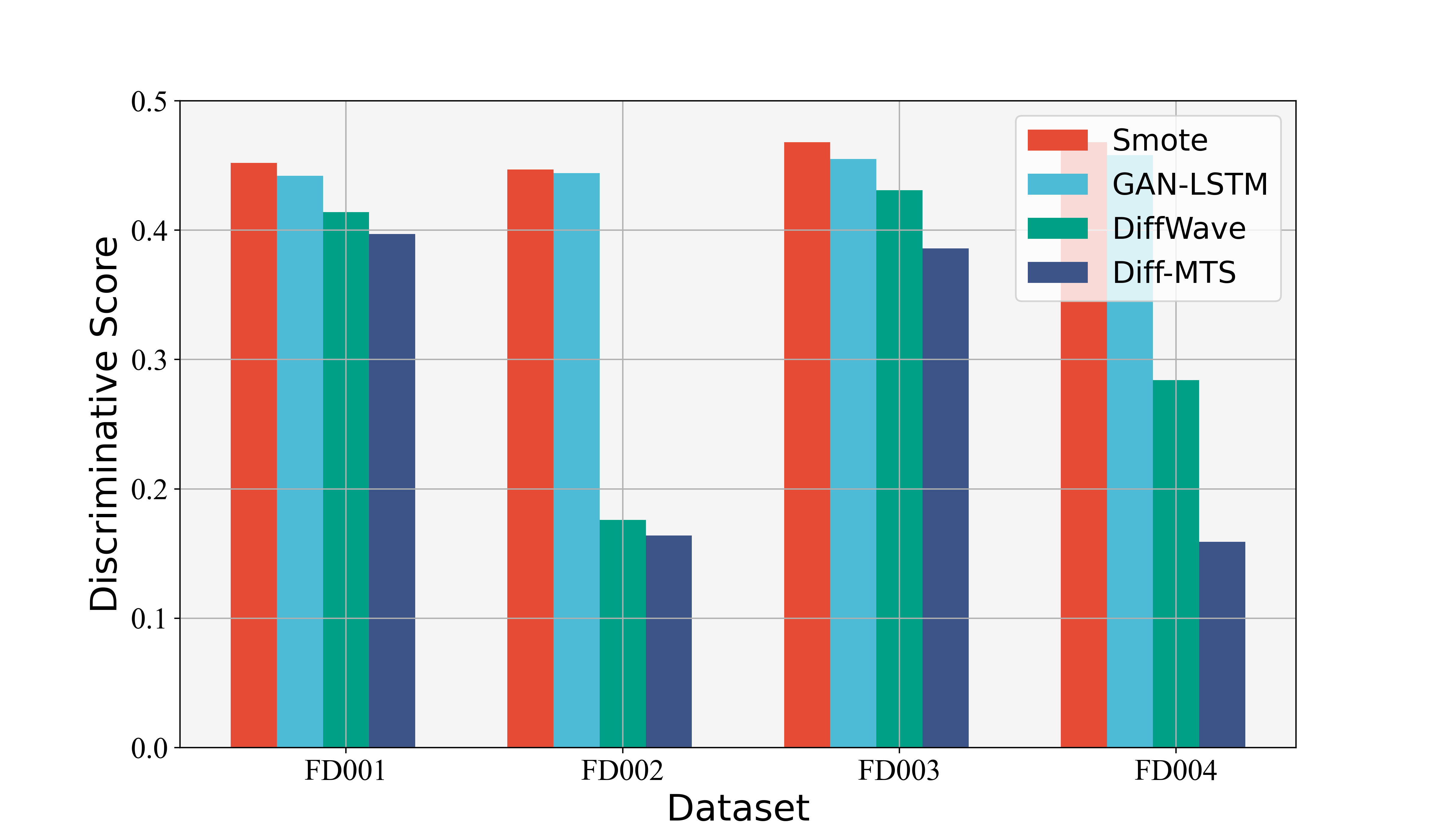}}
	\hfill    
	\caption{Discriminative scores of different generative methods across four datasets. Lower scores indicate higher fidelity.}
	\label{fig:plot_ds}
\end{figure}

\textbf{Fidelity of the synthetic data:} To evaluate the fidelity of generated time series, we compare Diff-MTS with diffusion-based methods (DiffWave\cite{kong2020diffwave}) and GAN-based methods (GAN-LSTM\cite{lu2021deep}) and SMOTE\cite{chawla2002smote}. Synthetic data is generated with lengths of 48, and fidelity is measured using the discriminative score, where lower values indicate higher similarity to real data. Fig. \ref{fig:plot_ds} illustrates the discriminative scores of different methods. Diff-MTS consistently achieves lower scores than GAN-based methods and SMOTE, demonstrating superior fidelity. For the four datasets, Diff-MTS achieves discriminative scores of 0.397, 0.164, 0.386, and 0.159, respectively, outperforming traditional generative methods SMOTE of 0.452, 0.447, 0.468, and 0.468. These results further demonstrate the superior fidelity of Diff-MTS in generating synthetic time series with a closer match to the real data distributions.

\begin{figure}[htpb]
	\centering
    \subfigure[Diffusion-based method]{
        \includegraphics[width=0.45\linewidth]{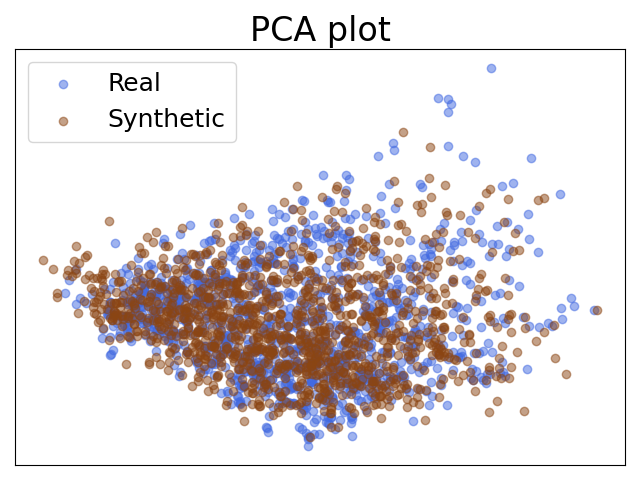}
        \includegraphics[width=0.45\linewidth]{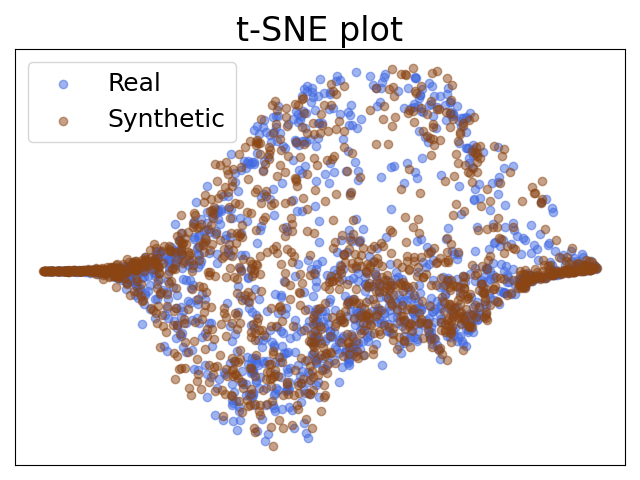}
        }
    \hspace{0in}  
    \subfigure[GAN-based method]{
        \includegraphics[width=0.45\linewidth]{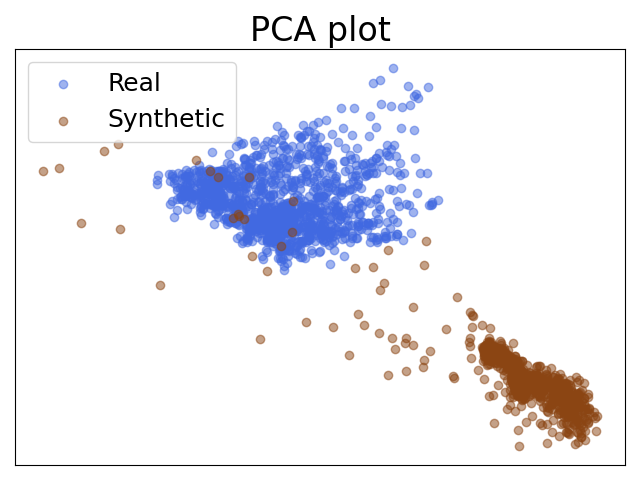}
        \includegraphics[width=0.45\linewidth]{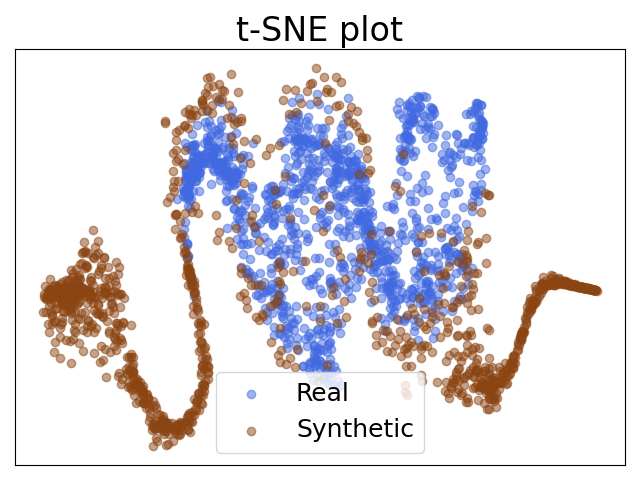}}
    \hspace{0in}  
    \subfigure[Traditional method]{
        \includegraphics[width=0.45\linewidth]{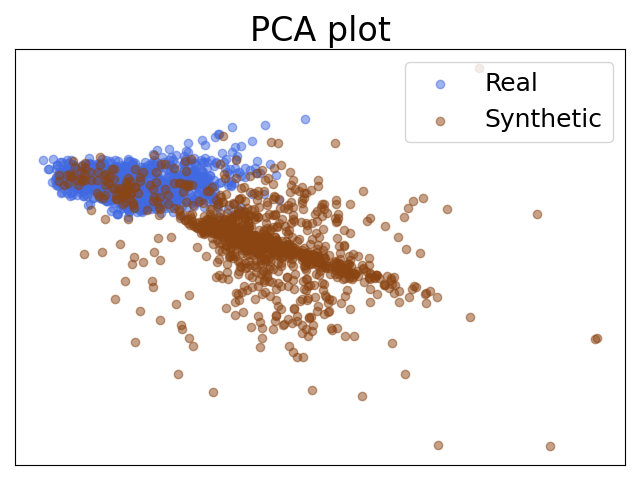}
        \includegraphics[width=0.45\linewidth]{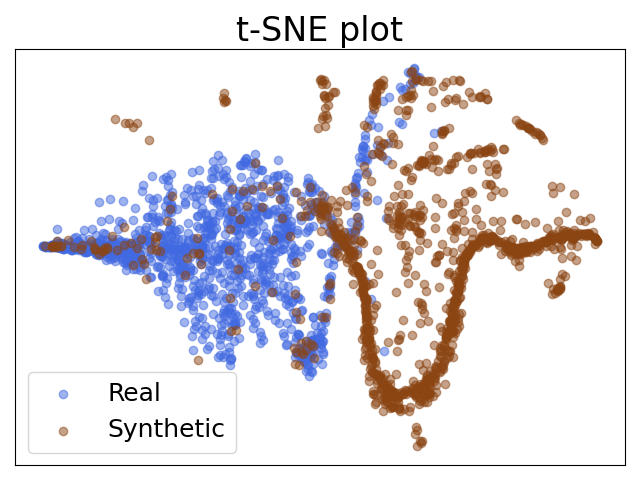}}
    \hspace{0in}  

    \caption{The synthetic and original data are visualized using PCA (first column) and t-SNE (second column). Each row displays both visualizations for the three methods. Original samples are depicted in blue, while synthesized samples are shown in brown. Greater overlap between points indicates higher similarity between the two data types.}
    \label{fig:pca}
\end{figure}

\textbf{Diversity of the synthetic data:} To assess the diversity of synthetic data, we apply PCA and t-SNE to reduce the dimensionality of real and generated time series data. We compare diffusion-based methods (Diff-MTS) with GAN-based methods and traditional method (SMOTE) by visualizing the distributions in two-dimensional space. Fig. \ref{fig:pca} presents the PCA and t-SNE visualizations for different methods. Diff-MTS exhibits broad and well-mixed distributions with real data, indicating high diversity. In contrast, GAN-based methods show noticeable clustering and partial overlap, suggesting limited variability. SMOTE-generated data remain highly concentrated, demonstrating the least diversity. These results confirm that the diffusion-based method generates more diverse synthetic data, effectively capturing the variability of industrial time series.

\begin{table}[!htpb]
\centering
\caption{Performance comparison of predictive maintenance using different data augmentation strategies}
\label{tab:data augmentation}
\begin{tabular}{ccccccc}
\hline
\multirow{2}{*}{Method}    & Data                  & \multirow{2}{*}{Metrics} & \multirow{2}{*}{FD001} & \multirow{2}{*}{FD002} & \multirow{2}{*}{FD003} & \multirow{2}{*}{FD004} \\
                           & mixture               &                          &                        &                        &                        &                        \\ \hline
Original                   & \multirow{2}{*}{} & RMSE                     & 14.453                 & 22.087                 & 14.488                 & 27.971                 \\
dataset                    &                       & MAE                      & 10.617                 & 17.033                 & 10.686                 & 22.305                 \\ \hline
\multirow{2}{*}{SMOTE}     & \multirow{2}{*}{\XSolidBrush} & RMSE                     & 23.647                 & 27.481                 & 27.231                 & 34.430                 \\
                           &                       & MAE                      & 17.241                 & 22.960                 & 21.851                 & 28.980                 \\
\multirow{2}{*}{Diffusion} & \multirow{2}{*}{\XSolidBrush} & RMSE                     & 13.688                 & 23.511                 & 13.495                 & 19.421                 \\
                           &                       & MAE                      & 9.846                  & 19.539                 & 9.564                  & 15.350                 \\ \hline
\multirow{2}{*}{SMOTE}     & \multirow{2}{*}{\checkmark} & RMSE                     & 13.585                 & 19.038                 & 16.683                 & 16.493                 \\
                           &                       & MAE                      & 10.149                 & 16.305                 & 12.581                 & 12.815                 \\
\multirow{2}{*}{Diffusion} & \multirow{2}{*}{\checkmark} & \textbf{RMSE}            & \textbf{13.446}        & \textbf{18.341}        & \textbf{13.412}        & \textbf{16.451}        \\
                           &                       & \textbf{MAE}             & \textbf{9.387}         & \textbf{15.295}        & \textbf{9.921}         & \textbf{12.615}        \\ \hline
\end{tabular}
\end{table}

\textbf{Performance of downstream application using limited data augmentation:}
This experiment leverages an industrial time-series diffusion model (Diff-MTS) to generate synthetic sensor data at three times the volume of the original dataset, which is then combined with the original data to enhance the training set for predictive maintenance tasks. As listed in Table \ref{tab:data augmentation}, comparative methods include the classic oversampling technique SMOTE and a baseline model using only the original data. The evaluation is conducted on aircraft engine sensor datasets (FD001-FD004), with Root Mean Squared Error (RMSE) and Mean Absolute Error (MAE) as the performance metrics. Here, "Data mixture" denotes the training strategy that combines synthetic and original data, whereas the absence of mixing indicates the use of synthetic data alone.

The results demonstrate that the synthetic data generated by the diffusion model significantly improves model performance. Under the mixed training strategy, the diffusion method achieves optimal results across all four subsets. For FD001, RMSE is 13.446 (a 7.0\% improvement over the original data) and MAE is 9.387 (an 11.6\% improvement); for FD004, RMSE and MAE are reduced to 16.451 (a 41.2\% improvement) and 12.615 (a 43.4\% improvement), respectively. Compared to the SMOTE mixed strategy, the diffusion method outperform it consistently across FD001-FD004. Notably, on the FD004 dataset, which represents complex operating conditions, the diffusion method further reduces RMSE by 19.6\% compared to SMOTE, highlighting its superior capability in capturing non-stationary temporal patterns. Further analysis reveals that even when using only the synthetic data generated by the diffusion model (without mixing with the original data), the RMSE (13.688) and MAE (9.846) already outperform those of the original dataset (14.453 and 10.617) on the FD001, indicating the high fidelity of the synthetic data.

\color{black}

\section{Challenges, Future Directions, and Conclusion}

\subsection{Generative Models for Low-quality and Mixed-type Data}
Industrial time series data involves a wide variety of sensors and devices, and therefore a wide variety of data types. It includes continuous data, such as temperature and pressure readings from sensors. It also includes discrete data, such as equipment status and switching information. This diversity makes the types of industrial time series data complex and variable, requiring generative models with the ability to handle different types of data. At the same time, industrial scenes often have quality issues, such as noise and outliers due to equipment malfunctions. This can affect the modeling of the data and lead to a reduction in the fidelity of the generated data.

To address data quality issues, the collected data can be pre-processed before being input into the model. In addition, more robust generative models can be developed to effectively deal with quality issues of noise and outliers in the industry. For mixed data types, the generation of mixed data can be facilitated by co-representation. For instance, Lee et al.\cite{lee2023codi} proposed a co-evolving contrastive generation framework to separately model continuous and discrete variables of tabular data. In addition, a general framework can be developed to include the generation, estimation, and prediction of mixed-type time series models.

\subsection{Generative Models with Long Temporal Expressive Power}
Industrial time series generative models face critical challenges in dealing with long-term temporal dependencies. industrial time series data typically have strong temporal dependencies where current values are influenced by previous values, and models need to be able to effectively capture these long-term dependencies in order to generate data with a reasonable temporal structure. 

Future research could focus on exploring more advanced sequence modeling methods to handle long-term temporal dependencies more effectively. Advanced recurrent neural network (RNN) structure is one such avenue. In addition, the attention mechanism\cite{ren2023time} and state space models\cite{austin2021structured} are also effective methods.

\subsection{Interpretable/Reliable/Credible Generative Models}

Deep generative models face the challenges of lack of interpretability and unreliability of the generated data. Reliability requires that the data generated accurately reflects the true behavior of the industrial system and ensures that the model is producing trustworthy results. Interpretability requires that the generated data is not only accurate but also needs to be able to be interpreted and understood to meet the needs of domain experts in the modeling decision-making process.

Future research is directed towards generative models that incorporate physical information to meet the high demand for reliability and interpretability in industry. Xiong et al.\cite{47-xiong2023controlled} presents the Controlled Physics-Informed Generative Adversarial Network (CPI-GAN), a hybrid framework that
synthesizes degradation trajectories for enhancing remaining useful life predictions. This integrated approach is expected to improve the reliability and interpretability of the generated data and provide a more credible database for industrial applications. By incorporating physical information into the generative models, it can be ensured that the generated data is not only based on the results of statistical learning but also takes into account the real mechanisms within the system.

\subsection{Industrial Multimodal Generative Models}

Industrial systems often contain multiple sensors and data sources that may have different modalities such as text, numerical values, etc. The challenge is that models need to be able to effectively handle such multi-modal data. The challenge is that the model needs to be able to efficiently handle such multi-modal data, where the data structure and feature representation of each modality may differ significantly.
The data generated by different sensors may be heterogeneous, increasing the complexity of model fusion and correlation.

Future research directions should focus on the development of generative models applicable to multimodal data. This may involve an in-depth study of representation learning methods for multimodal data to better capture the correlations among different sensors, which will help the models to more fully simulate the behavior and state of industrial systems. In addition, by joining modalities from different data not only can performance be enhanced but possible tasks can be explored. For instance, AnomalyGPT\cite{gu2023anomalygpt} integrates the information of Industrial Anomaly Detection (IAD) task to the Large Vision-Language Models (LVLMs) to support the multi-dialogues. Moreover, it doesn’t need manual threshold adjustment, which can directly present the anomalies.  Multimodal time series generation can be performed by combining text and signals.

\subsection{Industrial Generalized Generative Models}

Industrial environments can have a diversity of processes, equipment configurations, and operating conditions, leading to the need for models that can adapt and maintain high performance in different industrial environments. Trained DGMs often lack generalization capabilities as they can only generate samples that are consistent with the data in the training dataset, but can not generate new data samples. 

To enable generative models to learn richer, generic feature representations, training large generative models on large-scale industrial time series datasets is a promising approach. There are a number of time series base models that take advantage of the powerful representational capabilities of large language models to enhance their time series modeling capabilities, such as One fits all\cite{zhou2024one}, LLM4TS\cite{chang2023llm4ts}. For example, One fits all\cite {zhou2024one} uses frozen pre-trained language models to attain state-of-the-art performance across a variety of time series tasks. Integrating large generative models like ChatGPT\cite{109-ouyang2022training} or specialized time series models further amplifies the adaptability of the generative model. This combination not only improves performance but also enables the generation of data across different domains.

\subsection{Generative Models for Smart Manufacturing}
Generative modeling can generate a variety of industrial design prototypes to help engineers explore diverse device concepts and accelerate product design and innovation, including code generation\cite{111-li2022competition}, production process generation, and so on. The advantages of DGMs include: (1) Generative design models can produce diverse designs and stimulate the creativity of designers. (2) Automated code generation can reduce the cost of generation and design, especially in mass production. For instance, the paper\cite{113-lu2020procedure2command} proposed a generative AI methodology that automatically translates nuclear power plant safety operation flowcharts into executable code to improve the efficiency and reduce the cognitive burden of operators using flowcharts.

\subsection{Ethical and Privacy Consideration in Generative Models}
The deployment of generative models may introduce critical ethical and privacy challenges. The first issue is algorithmic bias. When generative models are trained on historical industrial data, they can inadvertently inherit and even amplify existing biases. To address this, techniques such as fairness-aware training\cite{choi2024fair} and adversarial debiasing\cite{parihar2024balancing} have proven effective in reducing correlations between generated outputs and sensitive attributes, ensuring that the model does not perpetuate historical inequities. These approaches help recalibrate the model’s learning process, promoting a more balanced and just representation of the underlying industrial data.

The second issue is data privacy and security. Industrial datasets may contain sensitive, proprietary information, and the misuse or leakage of such data could have severe consequences. Integrating differential privacy into the training process can ensure that individual data points remain unidentifiable, even in the synthetic outputs generated by the model. Furthermore, adopting federated learning frameworks\cite{zhang2024modeling} allows multiple stakeholders to collaboratively train models without the need to exchange raw data, thereby preserving data locality and enhancing security.

\subsection{Conclusion}
Currently, LGMs and DGMs have received much attention due to their ability to understand, generate, and create data. For this reason, generative models have made significant progress and show great potential in industrial time series. In conclusion, this review provides a detailed overview of the generative models in industrial. First, we analyze the definition, characteristics, and data scarcity challenges within industrial time series generation. Then, we propose a DGM-based AIGC framework for industrial data generation and present a  multi-perspective categorization of industrial DGMs. We further discuss the application of DGMs in industrial time series. Additionally, we establish a comprehensive evaluation benchmark for generative models, systematically assessing fidelity, diversity, and utility. In addition, we explore the construction of LGMs in the industrial domain. Finally, we conclude with challenges and future directions to enable the development of generative models in industrial applications.

\color{black}
\bibliographystyle{ieeetr}
\bibliography{reference-abb}

\begin{IEEEbiography}[{\includegraphics[width=1in,height=2.5in,clip,keepaspectratio]{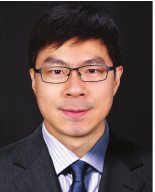}}]{Lei Ren} (Member, IEEE) received the Ph.D. degree in computer science from the Institute of Software, Chinese Academy of Sciences, Beijing, China, in 2009. 
    
He is currently a Professor with the School of Automation Science and Electrical Engineering, Beihang University, Beijing, China, and also with the Zhongguancun Laboratory, Beijing, China. His research interests include neural networks and deep learning, time series analysis, and industrial AI applications. Dr. Ren serves as an Associate Editor for the IEEE TRANSACTIONS ON NEURAL NETWORKS AND LEARNING SYSTEMS and other international journals.	
\end{IEEEbiography}

\begin{IEEEbiography}[{\includegraphics[width=1in,height=2.5in,clip,keepaspectratio]{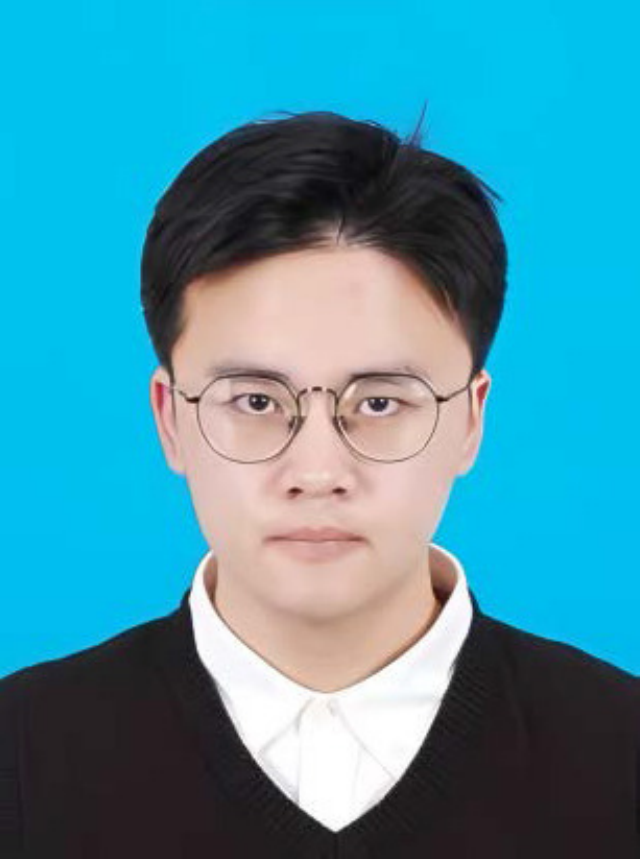}}]{Haiteng Wang} (Student Member, IEEE) received the B.Eng. Degree in automation engineering from Beihang University, Beijing, China, in 2021, where he is currently pursuing the Ph.D. degree with the School of Automation Science and Electrical Engineering. His current research interests include industrial time series and generative AI.
\end{IEEEbiography}

\begin{IEEEbiography}
[{\includegraphics[width=1in,height=2.5in,clip,keepaspectratio]{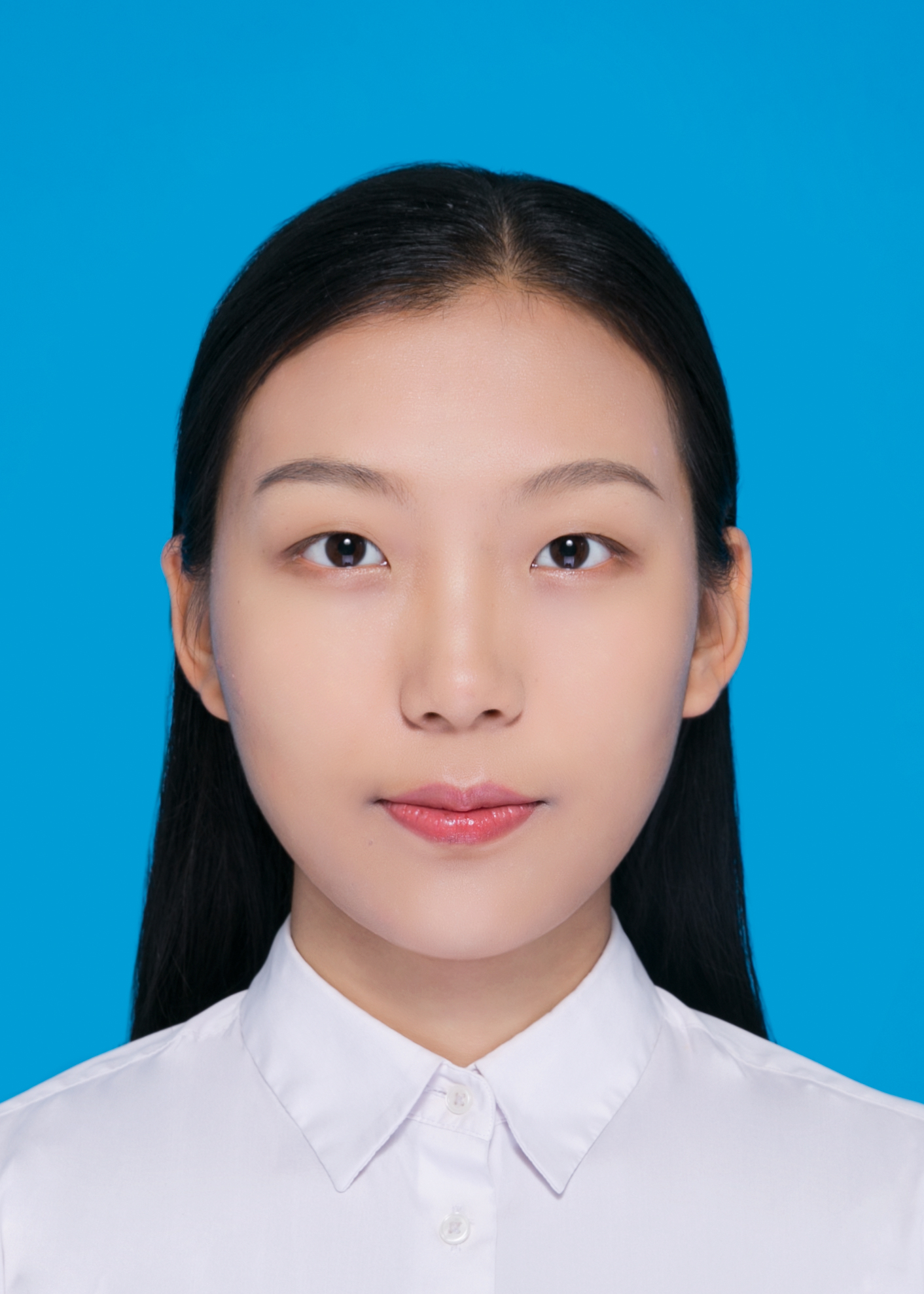}}]{Jinwang Li} (Student Member, IEEE) received her B.S. Degree from Beihang University, Beijing, China, in 2022 and is currently pursuing a Master’s degree. Her current research interest is Generative Artificial Intelligence.
\end{IEEEbiography}

\begin{IEEEbiography}[{\includegraphics[width=1in,height=2.5in,clip,keepaspectratio]{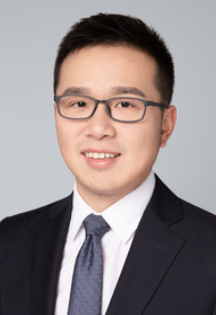}}]{Yang Tang} (Fellow, IEEE) received the B.S. and Ph.D. degrees in electrical engineering from Donghua University, Shanghai, China, in 2006 and 2010, respectively. 

From 2008 to 2010, he was a Research Associate with the Hong Kong Polytechnic University, Hong Kong. From 2011 to 2015, he was a Postdoctoral Researcher with the Humboldt University of Berlin, Berlin, Germany, and with the Potsdam Institute for Climate Impact Research, Potsdam, Germany. He is currently a Professor with the East China University of Science and Technology, Shanghai. He has authored or coauthored more than 200 papers in international journals and conferences, including more than 100 papers in IEEE Transactions and 20 papers in IFAC journals. His current research interests include distributed estimation/control/optimization, computer vision, reinforcement learning, cyberphysical systems, hybrid dynamical systems, and their applications. He was the recipient of the Alexander von Humboldt Fellowship. He is an Associate Editor of IEEE TRANSACTIONS ON NEURAL NETWORKS AND LEARNING SYSTEMS, IEEE TRANSACTIONS ON CYBERNETICS, IEEE TRANSACTIONS ON INDUSTRIAL INFORMATICS, IEEE/ASME TRANSACTIONS ON MECHATRONICS, IEEE TRANSACTIONS ON CIRCUITS AND SYSTEMS-I: REGULAR PAPERS, IEEE TRANSACTIONS ON COGNITIVE AND DEVELOPMENTAL SYSTEMS, IEEE TRANSACTIONS ON EMERGING TOPICS IN COMPUTATIONAL INTELLIGENCE, IEEE SYSTEMS JOURNAL, Engineering Applications of Artificial Intelligence (IFAC Journal) and Science China Information Sciences. He has been awarded as best/outstanding Associate Editor in IEEE journals four times. He is a (leading) Guest Editor for several special issues focusing on autonomous systems, robotics, and industrial intelligence in IEEE Transactions.
\end{IEEEbiography}

\begin{IEEEbiography}[{\includegraphics[width=1in,height=2.5in,clip,keepaspectratio]{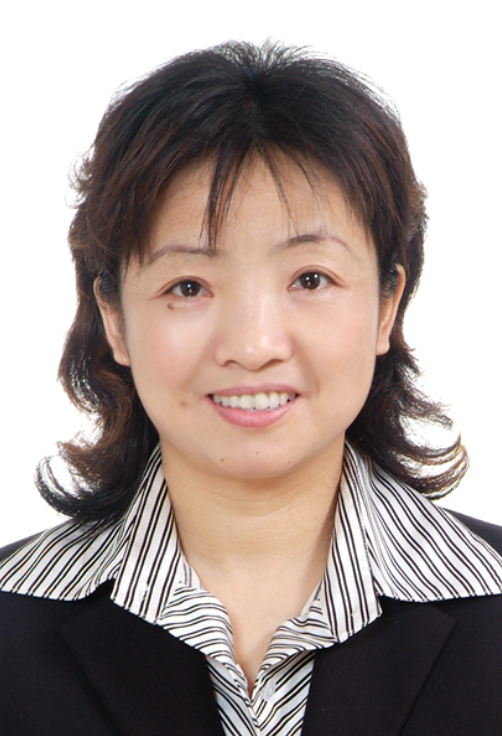}}]{Chunhua Yang} (Fellow, IEEE) received the M.S. degree in automatic control engineering and the Ph.D. degree in control science and engineering from Central South University, Changsha, China, in 1988 and 2002, respectively. 

Since 1999, she has been a Full Professor with the School of Information Science and Engineering, Central South University, where she is currently the HoD of the School of Automation.  Her current research interests include modeling and optimal control of complex industrial processes.

Dr. Yang is currently an Associate Editor for a number of journals, including IEEE TRANSACTIONS ON INDUSTRIAL ELECTRONICS and IEEE/ASME TRANSACTIONS ON MECHATRONICS.
\end{IEEEbiography}

\end{document}